\documentclass[dvipsnames]{article}
\usepackage{graphicx}
\usepackage{booktabs}
\usepackage{caption}
\usepackage{subcaption}
\usepackage{showlabels}
\usepackage{enumitem}
\usepackage{textcomp}
\usepackage[table,xcdraw]{xcolor}
\usepackage{overpic}
\usepackage{units}
\usepackage{tablefootnote}
\usepackage{multirow}

\usepackage{amsmath,amsfonts,bm}

\def\Figref#1{Figure~\ref{#1}}
\def\Tabref#1{Table~\ref{#1}}

\def\Eqref#1{Equation~\ref{#1}}
\def\eqref#1{\Eqref{#1}}

\def\1{\bm{1}}

\DeclareMathAlphabet{\mathsfit}{\encodingdefault}{\sfdefault}{m}{sl}
\SetMathAlphabet{\mathsfit}{bold}{\encodingdefault}{\sfdefault}{bx}{n}

\usepackage{subcaption}
\usepackage{bbm}
\usepackage{mathtools}
\usepackage{amsthm, amssymb}

\def\1{\bm{1}}

\newcommand{\norm}[1]{\lVert #1 \rVert}

\newcommand{\pp}[1]{\left( #1 \right)}

\usepackage{hyperref}
\hypersetup{
    unicode=false,
    pdftoolbar=true,
    pdfmenubar=true,
    pdffitwindow=false,
    pdfstartview={FitH},
    pdfauthor={Anonymous},
    pdfkeywords={Infinite Neural Networks, Neural Tangent Kernel, Gaussian Processes, JAX, Empirical Study},
    pdfnewwindow=true,
    colorlinks=True,
    linkcolor=BurntOrange,
    citecolor=RawSienna,
    filecolor=magenta,
    urlcolor=blue
}

\newcommand{\sref}[1]{\S\ref{#1}}

\PassOptionsToPackage{numbers, sort&compress}{natbib}

\usepackage[preprint]{neurips_2020}

\usepackage[utf8]{inputenc} %
\usepackage[T1]{fontenc}    %
\usepackage{hyperref}       %
\usepackage{url}            %
\usepackage{booktabs}       %
\usepackage{amsfonts}       %
\usepackage{nicefrac}       %
\usepackage{microtype}      %
\usepackage{hyperref}

\newcommand*\samethanks[1][\value{footnote}]{\color{orange}\footnotemark[#1]}

\definecolor{darkgreen}{rgb}{0,0.6,0}
\newcommand{\jl}[1]{{\color{red}[JL: #1]}}
\newcommand{\jp}[1]{{\color{purple}[JP: #1]}}
\newcommand{\js}[1]{{\color{darkgreen}[JSD: #1]}}
\newcommand{\rcom}[1]{{\color{orange}[RN: #1]}}
\newcommand{\xcom}[1]{{\color{blue}[XL: #1]}}
\newcommand{\scom}[1]{{\color{brown}[SS: #1]}}

\renewcommand{\jl}[1]{}
\renewcommand{\jp}[1]{}
\renewcommand{\js}[1]{}
\renewcommand{\rcom}[1]{}
\renewcommand{\xcom}[1]{}
\renewcommand{\scom}[1]{}

\newcommand{\capsize}{}

\usepackage[compact]{titlesec}

\title{Finite Versus Infinite Neural Networks:\\ an Empirical Study}

\newcommand{\email}[1]{\tt\small\href{mailto:#1@google.com}{#1}}

\author{%
  Jaehoon Lee\\
\And
  Samuel S. Schoenholz\thanks{SSS, JP, BA, LX, and RN contributed equally.}\\
\And 
  Jeffrey Pennington\samethanks[1]\\
\And 
  Ben Adlam\samethanks[1]\, \thanks{Work done as a member of the Google AI Residency program (\href{https://g.co/airesidency}{https://g.co/airesidency}).}\\
\AND 
  Lechao Xiao\samethanks[1]\\
\And 
  Roman Novak\samethanks[1]\\
\And 
  Jascha Sohl-Dickstein\\
\AND
{}\vspace{-0.7cm}\\
Google Brain\\
\texttt{\{\email{jaehlee}, \email{schsam}, \email{jpennin}, \email{adlam}, \email{xlc}, \email{romann}, \email{jaschasd}\}@google.com}
}

\begin{document}

\maketitle

\begin{abstract}

We perform a careful, thorough, and large scale empirical study of the correspondence between wide neural networks and kernel methods. By doing so, we resolve a variety of open questions related to the study of infinitely wide neural networks. Our experimental results include: kernel methods outperform fully-connected finite-width networks, but underperform convolutional finite width networks; neural network Gaussian process (NNGP) kernels frequently outperform neural tangent (NT) kernels; 
centered and ensembled finite networks have reduced posterior variance and behave more similarly to infinite networks; 
weight decay and the use of a large learning rate break the correspondence between finite and infinite networks; the NTK parameterization outperforms the standard parameterization for finite width networks; 
diagonal regularization of kernels acts similarly to early stopping; 
floating point precision limits kernel performance beyond a critical dataset size; 
regularized ZCA whitening improves accuracy;
finite network performance depends non-monotonically on width in ways not captured by double descent phenomena;
equivariance of CNNs is only beneficial for narrow networks far from the kernel regime. 
Our experiments additionally motivate an improved layer-wise scaling for weight decay which improves generalization in finite-width networks. Finally, we develop improved best practices for using NNGP and NT kernels for prediction, including a novel ensembling technique. Using these best practices we achieve state-of-the-art results on CIFAR-10 classification for kernels corresponding to each architecture class we consider.

\end{abstract}

\section{Introduction}

A broad class of both Bayesian 
\citep{neal, williams1997, hazan2015steps, lee2018deep, matthews2018, matthews2018b_arxiv, Borovykh2018, garriga2018deep, novak2018bayesian, yang2017mean, yang2018a, pretorius2019expected, yang2019scaling, yang2019wide, neuraltangents2020, hron2020, Hu2020InfinitelyWG} and gradient descent trained \citep{Jacot2018ntk, li2018learning, allen2018convergence, du2018gradient, du2018gradienta, zou2018stochastic, lee2019wide, chizat2019lazy, arora2019on, sohl2020infinite, Huang2020OnTN, du2019graph, yang2019scaling, yang2019wide, neuraltangents2020, hron2020} neural networks converge to Gaussian Processes (GPs) or closely-related kernel methods as their intermediate layers are made infinitely wide. 
The predictions of these infinite width networks are described by the Neural Network Gaussian Process (NNGP) \citep{lee2018deep,matthews2018} kernel for Bayesian networks, and by the Neural Tangent Kernel (NTK) \citep{Jacot2018ntk} and weight space linearization \citep{lee2019wide,chizat2019lazy} for gradient descent trained networks. 

This correspondence has been key to recent breakthroughs in our understanding of neural networks \citep{xiao18a, valle-perez2018deep, wei2019regularization, xiao2019disentangling, NIPS2019_9449, ben2019role, yang2019fine, ober2020global, Hu2020Provable, lewkowycz2020large, lewkowycz2020training}. 
It has also enabled practical advances in kernel methods \citep{garriga2018deep, novak2018bayesian, arora2019on, li2019enhanced, Arora2020Harnessing, Shankar2020NeuralKW, neuraltangents2020, hron2020}, Bayesian deep learning \citep{wang2018function, Cheng_2019_CVPR, Carvalho2020ScalableUF}, active learning \citep{ijcai2019-499}, and semi-supervised learning \citep{Hu2020InfinitelyWG}.
The NNGP, NTK, and related large width limits \citep{cho2009kernel, daniely2016toward, poole2016exponential, chen2018rnn, li2018on, daniely2017sgd,  pretorius2018critical, hayou2018selection, karakida2018universal, blumenfeld2019mean, hayou2019meanfield, schoenholz2016deep, pennington2017resurrecting, xiao18a, yang2017mean, geiger2019disentangling, geiger2020scaling, antognini2019finite,  Dyer2020Asymptotics, huang2019dynamics, yaida2019non} are unique in giving an exact theoretical description of large scale neural networks.
Because of this, we believe they will continue to play a transformative role in deep learning theory.

Infinite networks are a newly active field, and 
foundational
empirical questions remain unanswered. 
In this work, we perform an extensive and in-depth empirical study of finite and infinite width neural networks. 
In so doing, we provide quantitative answers to questions about the factors of variation that drive performance in finite networks and kernel methods, uncover surprising new behaviors, and develop best practices that improve the performance of both finite and infinite width networks.
We believe our results will both ground and motivate future work in wide networks.

\section{Experiment design}\label{sec:experimental_design}

\definecolor{ntk_param}{RGB}{255, 140, 0}
\definecolor{standard_param}{RGB}{65, 105, 225}

\vspace{-0.15cm}\begin{figure}[t]
\hfill\hfill Finite gradient descent (GD) \hfill \quad\quad Infinite GD \quad Infinite Bayesian
\vspace{-0.25cm}
\centering
\includegraphics[width=1.02\columnwidth]{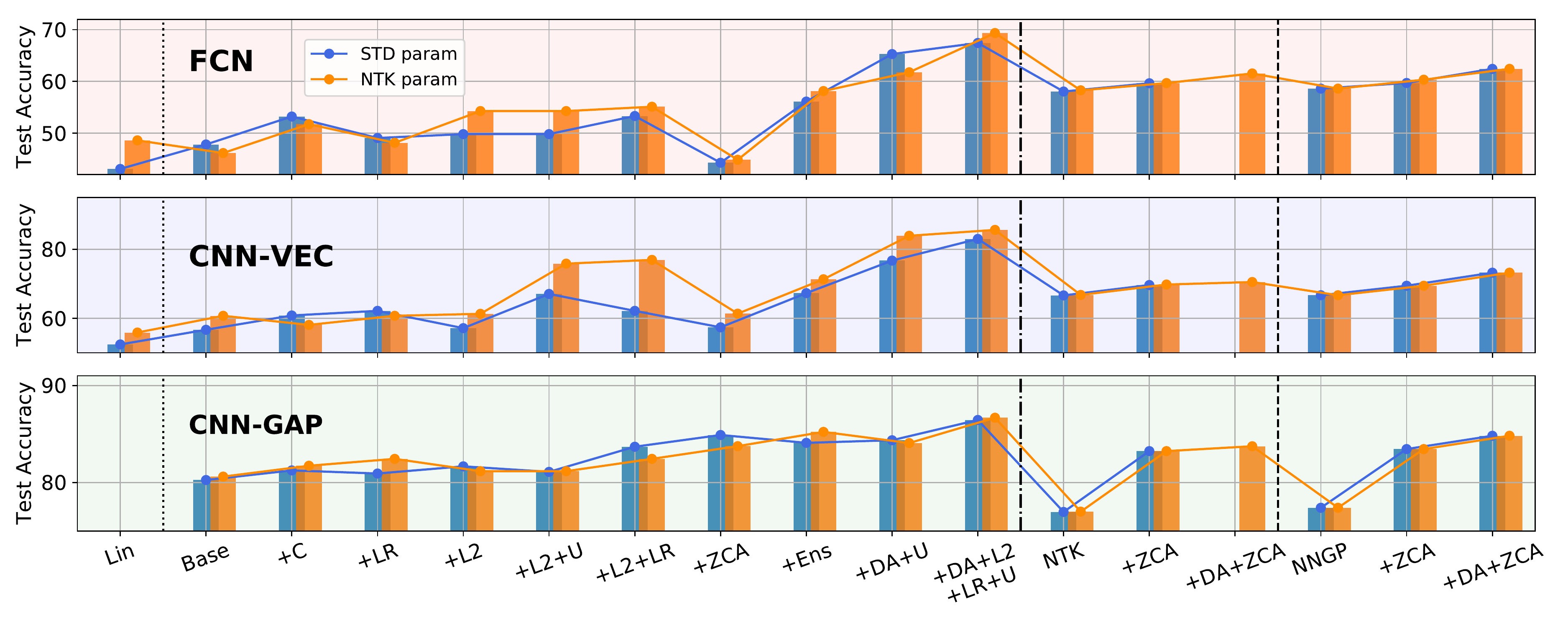}
\caption{
\capsize
\textbf{CIFAR-10 test accuracy for finite and infinite networks and their variations}. Starting from the 
finite width %
\texttt{base} network of given architecture class described in \sref{sec:experimental_design}, performance changes from \textbf{centering} (\texttt{+C}), \textbf{large learning rate} (\texttt{+LR}), allowing \textbf{underfitting} by early stopping (\texttt{+U}), input preprocessing with \textbf{ZCA regularization} (\texttt{+ZCA}), multiple initialization \textbf{ensembling} (\texttt{+Ens}), and some combinations are shown, for {\color{standard_param}\textbf{Standard}} and {\color{ntk_param}\textbf{NTK}} parameterizations. 
The performance of the \textbf{linearized} (\texttt{lin}) base network is also shown. 
See \Tabref{tab:main-table} for precise values for each of these experiments, as well as for additional experimental conditions not shown here.}
\label{fig:tricks_vs_accuracy}
\end{figure}

To systematically develop a phenomenology of infinite and finite neural networks, we first establish base cases for each architecture where infinite-width kernel methods, linearized weight-space networks, and nonlinear gradient descent based training can be directly compared. In the finite-width settings, the base case uses 
mini-batch gradient descent at a constant small learning rate~\cite{lee2019wide} with MSE loss (implementation details in~\sref{app:batch-size}). In the kernel-learning setting we compute the NNGP and NTK for the entire dataset and do exact inference as described in~\cite[page 16]{rasmussen2006gaussian}. Once this one-to-one comparison has been established, we augment the base setting with a wide range of interventions. We discuss each of these interventions in detail below. Some interventions will approximately preserve the correspondence (for example, data augmentation), while others explicitly break the correspondence in a way that has been hypothesized in the literature to affect performance (for example, large learning rates~\cite{lewkowycz2020large}). 
We additionally explore linearizing the base model around its initialization, in which case its training dynamics become exactly described by a constant kernel. This differs from the kernel setting described above due to finite width effects.

We use MSE loss to allow for easier comparison to kernel methods, whose predictions can be evaluated in closed form for MSE. 
See \Tabref{tab:xent-vs-mse} and \Figref{fig:xent-vs-mse} for a comparison of MSE to softmax-cross-entropy loss. 
Softmax-cross-entropy provides a consistent small benefit over MSE, and will be interesting to consider in future work.

Architectures we work with are built from either Fully-Connected (\texttt{FCN})
or Convolutional (\texttt{CNN}) layers. In all cases we use ReLU nonlinearities with critical initialization with small bias variance ($\sigma_w^2=2.0, \sigma_b^2=0.01$). Except if otherwise stated, we consider \texttt{FCN}s with 3-layers of width 2048 and \texttt{CNN}s with 8-layers of 512 channels per layer. For convolutional networks we must collapse the spatial dimensions of image-shaped data before the final readout layer. To do this we either: flatten the image into a one-dimensional vector (\texttt{VEC}) or apply global average pooling to the spatial dimensions (\texttt{GAP}). 
Finally, we compare two ways of parameterizing the weights and biases of the network: the standard parameterization (STD), which is used in work on finite-width networks, and the NTK parameterization (NTK) which has been 
used in 
most infinite-width studies to date (see~\cite{sohl2020infinite} for the standard parameterization at infinite width).

Except where noted, for all kernel experiments we optimize over diagonal kernel regularization
independently for each experiment.
For finite width networks, except where noted we use a small learning rate corresponding to the base case. See \sref{app hyperparameters} for details.

The experiments described in this paper are often very compute intensive. For example, to compute the NTK or NNGP for the entirety of CIFAR-10 for \texttt{CNN-GAP} architectures one must explicitly evaluate the entries in a $6\times10^7$-by-$6\times10^7$ 
kernel matrix. Typically this takes around 1200 GPU hours with double precision, and so we implement our experiments via massively distributed compute infrastructure based on beam~\cite{beam}. All experiments use the Neural Tangents library~\cite{neuraltangents2020}, built on top of JAX~\cite{jax2018github}. 

To be as systematic as possible while also tractable given this large computational requirement, we evaluated every intervention for every architecture and focused on a single dataset, CIFAR-10~\cite{krizhevsky2009learning}. However, to ensure robustness of our results across dataset, we evaluate several key claims on CIFAR-100 and Fashion-MNIST~\cite{xiao2017/online}.

\section{Observed empirical phenomena}

\subsection{NNGP/NTK can outperform finite networks}
\label{sec:infinite-vs-finite}
A common 
assumption
in the study of infinite networks is that they underperform the corresponding finite network in the large data regime.
We carefully examine this assumption, by comparing kernel methods against the base case of a finite width architecture trained with small learning rate and no regularization (\sref{sec:experimental_design}), and then individually examining the effects of common training practices which break (large LR, L2 regularization) or improve (ensembling) the infinite width correspondence to kernel methods.
The results of these experiments are 
summarized in \Figref{fig:tricks_vs_accuracy} and \Tabref{tab:main-table}.

First focusing on base finite networks, we observe that infinite \texttt{FCN} and \texttt{CNN-VEC} outperform their respective finite networks. On the other hand, infinite \texttt{CNN-GAP} networks perform worse than their finite-width counterparts in the base case, consistent with observations in~\citet{arora2019on}. We emphasize that architecture plays a key role in relative performance, in line with an observation made in \citet{geiger2019disentangling} in the study of lazy training. For example, infinite-\texttt{FCN}s outperform finite-width networks even when combined with various tricks such as high learning rate, L2, and underfitting. Here the performance becomes similar only after ensembling (\sref{sec:ensemble_of_networks}). 

One interesting observation is that ZCA regularization preprocessing (\sref{sec:zca}) can provide significant improvements to the \texttt{CNN-GAP} kernel, closing the gap to within 1-2\%. 

\subsection{NNGP typically outperforms NTK}
\label{sec:nngp-vs-ntk}
\begin{figure}
\centering
\includegraphics[width=\columnwidth]{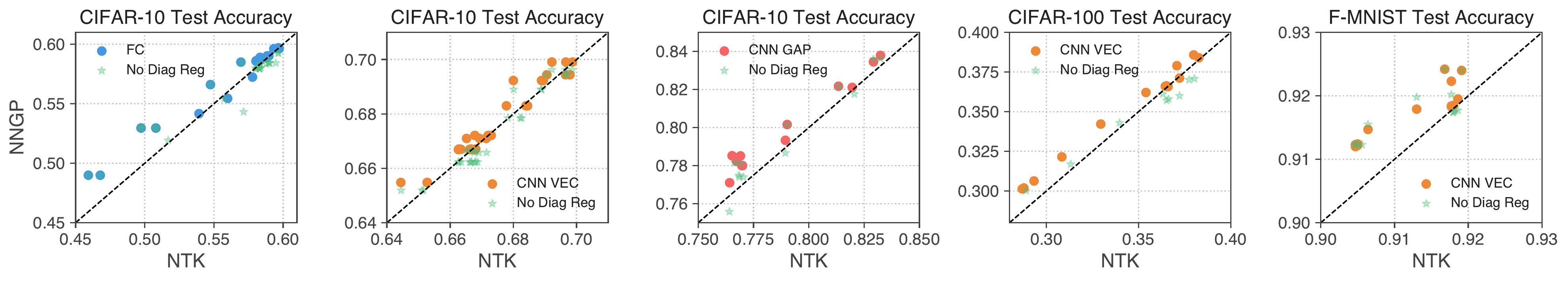}
\caption{
\capsize
\textbf{NNGP often outperforms NTK in image classification tasks when diagonal regularization is carefully tuned.} 
The performance of the NNGP and NT kernels are plotted against each other 
for a variety of data pre-processing configurations (\sref{sec:zca}),
while regularization (\Figref{fig reg compare}) is independently tuned for each.
}
\label{fig:nngp-vs-ntk}
\end{figure}

Recent %
evaluations of infinite width networks have put significant emphasis on the NTK, without explicit comparison against the respective NNGP models \citep{arora2019on, li2019enhanced, du2019graph, Arora2020Harnessing}. Combined with the view of NNGPs as ``weakly-trained'' \citep{lee2019wide, arora2019on} (i.e. having only the last layer learned), one might expect NTK to be a more effective model class than NNGP.
On the contrary, we usually
observe that NNGP inference achieves better performance. This can be seen in \Tabref{tab:main-table} where SOTA performance among fixed kernels is attained with the NNGP across all architectures. 
In \Figref{fig:nngp-vs-ntk} we show that this trend persists across CIFAR-10, CIFAR-100, and Fashion-MNIST (see~\Figref{fig:nngp-vs-ntk-uci} for similar trends on UCI regression tasks). 
In addition to producing stronger models, NNGP kernels require about half the memory and compute as the corresponding NTK, 
and some of the most performant kernels do not have an associated NTK at all~\cite{Shankar2020NeuralKW}. Together these results suggest that when approaching a new problem where the goal is to maximize performance, 
practitioners should start with the NNGP.

We emphasize that both tuning of the diagonal regularizer (\Figref{fig reg compare}) and sufficient numerical precision (\sref{sec:diag-reg}, \Figref{app kernel spectra}) were crucial to achieving an accurate comparison of these kernels.

\begin{figure}
\centering
\includegraphics[width=\columnwidth]{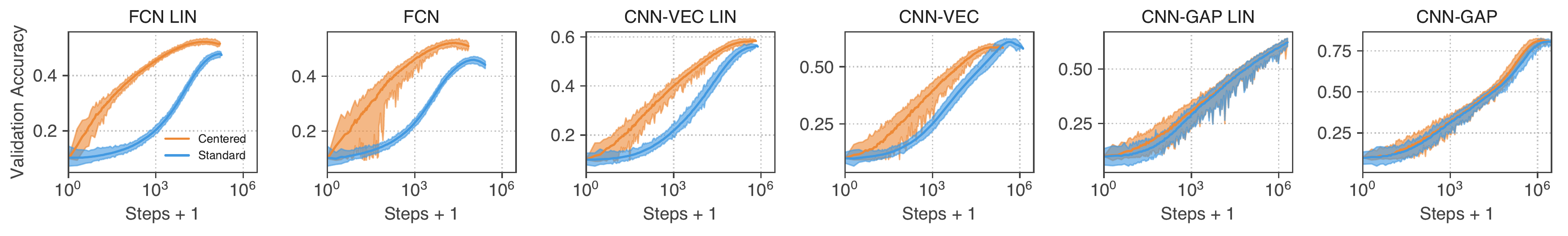}
\caption{
\capsize
\textbf{Centering can accelerate training and improve performance}. Validation accuracy throughout training for several finite width architectures. See \Figref{fig:training_curves} for training accuracy. 
}
\label{fig:validation_curves}
\end{figure}

\subsection{Centering and ensembling finite networks both lead to kernel-like performance}
\label{sec:ensemble_of_networks}
\begin{figure}
\centering
\includegraphics[width=\columnwidth]{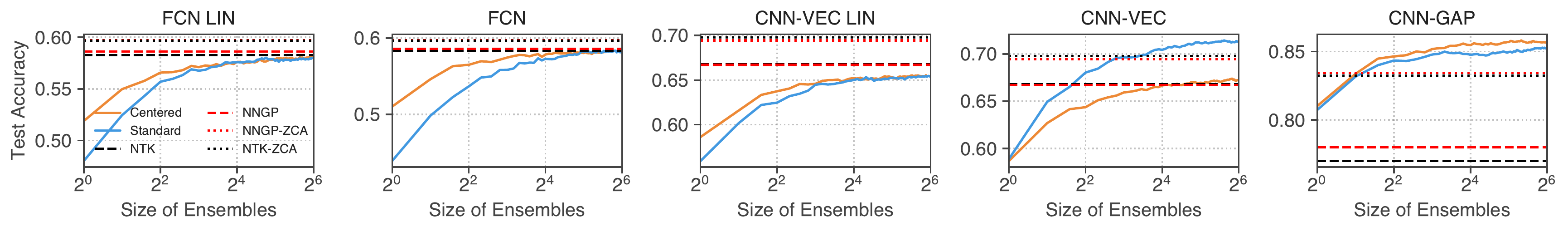}
\caption{
\capsize
\textbf{Ensembling base networks enables them to match the performance of kernel methods, and exceed kernel performance for nonlinear \texttt{CNN}s.} See \Figref{fig app ensemble} for test MSE.
}
\label{fig:ensemble}
\end{figure}

For overparameterized neural networks, some randomness from the initial parameters persists throughout training and the resulting learned functions are themselves random. This excess variance in the network's predictions generically increases the total test error through the variance term of the bias-variance decomposition. For infinite-width kernel systems this variance is eliminated by using the mean predictor. For finite-width models, the variance can be large, and test performance can be significantly improved by \emph{ensembling} a collection of models~\cite{geiger2019disentangling, geiger2020scaling}. %
In \Figref{fig:ensemble}, we examine the effect of ensembling. For \texttt{FCN}, ensembling closes the gap with kernel methods, suggesting that finite width \texttt{FCN}s underperform \texttt{FCN} kernels primarily due to variance.
For \texttt{CNN} models, ensembling also improves test performance, and ensembled \texttt{CNN-GAP} models significantly outperform the best kernel methods. 
The observation that ensembles of finite width \texttt{CNN}s can outperform infinite width networks while ensembles of finite \texttt{FCN}s cannot (see \Figref{fig:ensemble}) is consistent with earlier findings in~\cite{geiger2020scaling}.

Prediction variance can also be reduced by \emph{centering} the model, i.e. subtracting the model's initial predictions: $f_\text{centered}(t) = f(\theta(t)) - f(\theta(0))$. A similar variance reduction technique has been studied in~\cite{chizat2019lazy, zhang2019type, hu2020Simple, bai2020Beyond}. 
In \Figref{fig:validation_curves}, we observe that centering significantly speeds up training and improves generalization for \texttt{FCN} and \texttt{CNN-VEC} models, but has little-to-no effect on \texttt{CNN-GAP} architectures.
We observe that the scale posterior variance of \texttt{CNN-GAP}, in the infinite-width kernel, is small relative to the prior variance given more data, consistent with centering and ensembles having small effect.

\subsection{Large LRs and L2 regularization drive differences between finite networks and kernels}\label{sec:l2_lr}
In practice, L2 regularization (a.k.a. weight decay) or larger learning rates can break the correspondence between kernel methods and finite width neural network training even at large widths. 

\citet{lee2019wide} 
derives 
a critical learning rate $\eta_{\text{critical}}$ such that wide network training dynamics are equivalent to linearized training for $\eta< \eta_{\text{critical}}$.
\citet{lewkowycz2020large} argues that even at large width a learning rate $\eta \in (\eta_{\text{critical}}, c\cdot\eta_{\text{critical}})$ for a constant $c>1$ forces the network to move away from its initial high curvature minimum and converge to a lower curvature minimum, while \citet{li2019towards} argues that large initial learning rates enable networks to learn `hard-to-generalize' patterns.

In \Figref{fig:tricks_vs_accuracy} (and \Tabref{tab:main-table}), we observe that the effectiveness of a large learning rate (LR) is highly sensitive to both architecture and paramerization: LR improves performance of \texttt{FCN} and \texttt{CNN-GAP} by about $ 1\%$ for STD parameterization and about $2\%$ for NTK parameterization. In stark contrast, it has little effect on \texttt{CNN-VEC} with NTK parameterization and surprisingly, a huge performance boost on \texttt{CNN-VEC} with STD parameterization ($+5\%$).      

L2 regularization (\eqref{eq:l2-reg}) regularizes the squared distance between the parameters and the origin and encourages the network to converge to minima with smaller Euclidean norms. Such minima are different from those obtained by NT kernel-ridge regression (i.e. adding a diagonal regularization term to the NT kernel) \citep{wei2019regularization},
which essentially penalizes the deviation of the network's parameters from initialization \cite{hu2019understanding}.  See~\Figref{fig:reg-compare-sm} for a comparison.

L2 regularization consistently improves (+$1$-$2\%$) performance for all architectures and parameterizations. 
Even with a well-tuned L2 regularization, finite width \texttt{CNN-VEC} and \texttt{FCN} still underperform NNGP/NTK. 
Combining L2 with early stopping produces a dramatic additional $10\% - 15\%$ performance boost for finite width \texttt{CNN-VEC}, outperforming NNGP/NTK.
Finally, we note that L2+LR together provide a superlinear performance gain for all cases except \texttt{FCN} and \texttt{CNN-GAP} with NTK-parameterization. 
Understanding the nonlinear interactions between L2, LR, and early stopping on finite width networks is an important research question (e.g. see~\cite{lewkowycz2020large,lewkowycz2020training} for LR/L2 effect on the training dynamics). 

\subsection{Improving L2 regularization for networks using the standard parameterization}
\label{sec improved standard}

\begin{figure}
\centering
\begin{overpic}[width=\columnwidth]{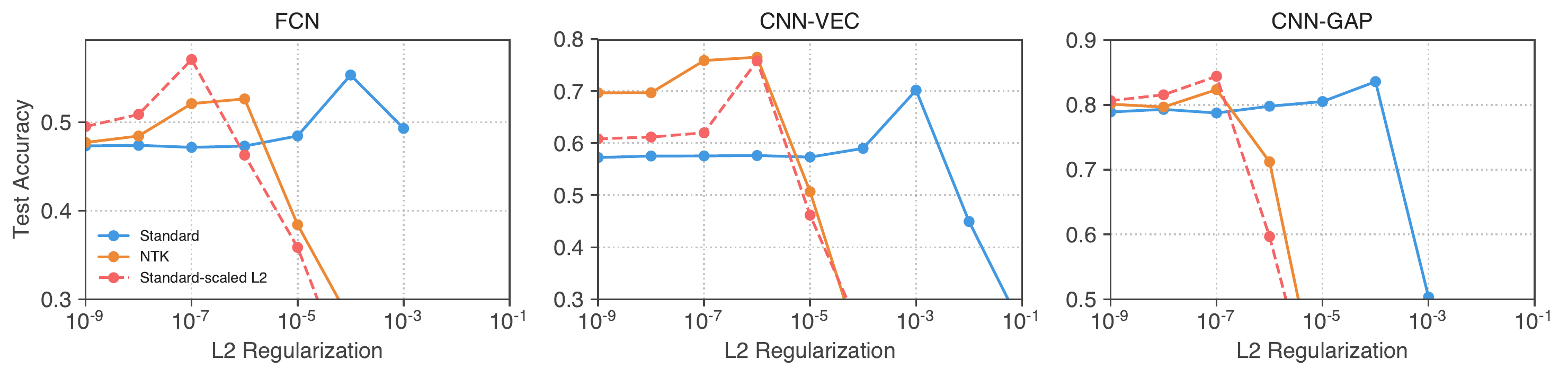}
\end{overpic}
\ \ 
\vspace{-0.45cm}
\caption{
\capsize
\textbf{Layerwise scaling motivated by NTK makes L2 regularization more helpful in standard parameterization networks.}
See \sref{sec improved standard} for introduction of the improved regularizer, \Figref{fig:l2-init} for further analysis on L2 regularization to initial weights, and \Figref{fig:reg-compare-sm} for effects on varying widths.
\label{fig reg compare}
}
\end{figure}

We find that L2 regularization provides dramatically more benefit (by up to $6\%$) to finite width networks with the NTK parameterization than to those that use the standard parameterization (see \Tabref{tab:main-table}). There is a bijective mapping between weights in networks with the two parameterizations, which preserves the function computed by both networks: $W^l_\text{STD} =  \nicefrac{W^l_\text{NTK}\,}{\sqrt{n^l}}$, where $W^l$ is the $l$th layer weight matrix, and $n^l$ is the width of the preceding activation vector. 
Motivated by the improved performance of the L2 regularizer in the NTK parameterization, we use this mapping to construct a regularizer for standard parameterization networks that produces the same penalty as vanilla L2 regularization would produce on the equivalent NTK-parameterized network. This modified regularizer is
    ${R}^{\text{STD}}_{\text{Layerwise}} = \frac{\lambda}{2} \sum_l n^l  \norm{W^l_\text{STD}}^2$.
This can be thought of as a layer-wise regularization constant $\lambda^l = \lambda n^l$. 
The improved performance of this regularizer is illustrated in \Figref{fig reg compare}.

\subsection{Performance can be non-monotonic in width beyond double descent}
\label{sec:perf_vs_width}

\begin{figure}
\centering
\includegraphics[width=\columnwidth]{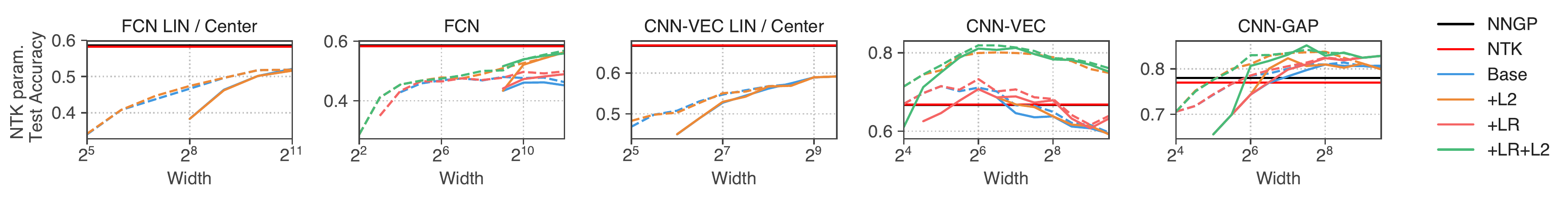}
\caption{
\capsize
\textbf{Finite width networks generally perform better with increasing width, but \texttt{CNN-VEC} shows surprising non-monotonic behavior.}
 {\bf L2}: non-zero weight decay allowed during training {\bf LR}: large learning rate allowed. Dashed lines are allowing underfitting (\textbf{U}). See \Figref{fig:width-combined} for plots for the standard parameterization, and \sref{sec:equivariance} for discussion of \texttt{CNN-VEC} results.
 }
\label{fig:width}
\end{figure}

Deep learning practitioners have repeatedly found that increasing the number of parameters in their models leads to improved performance~\citep{lawrence1998size, bartlett1998sample, Neyshabur2014InSO, canziani2016analysis, novak2018sensitivity, parkoptimal, novak2018bayesian}. 
While this behavior is consistent with a Bayesian perspective on generalization \citep{mackay1995probable,smith2017bayesian, wilson2020bayesian},
it seems at odds with classic generalization theory which primarily considers worst-case overfitting \citep{haussler1992decision, NIPS1988_154, Vapnik1998StatisticalLT, bartlett2002rademacher, bousquet2002stability, mukherjee2004statistical, poggio2004general}. This has led to a great deal of work on the interplay of overparameterization and generalization \citep{zhang2016understanding, advani2017high, neyshabur2018towards, neyshabur2018the, NIPS2018_8038, allen2019learning, ghorbani2019limitations, ghorbani2019linearized, arora2019fine, brutzkus19b}.
Of particular interest has been the phenomenon of double descent, in which performance increases overall with parameter account, but drops dramatically when the neural network is roughly critically parameterized~\citep{opper1990ability, belkin2019reconciling, nakkiran2019deep}.

Empirically, we find that in most cases (\texttt{FCN} and \texttt{CNN-GAP} in both parameterizations, \texttt{CNN-VEC} with standard parameterization) increasing width leads to monotonic improvements in performance. 
However, we also find a more complex dependence on width in specific relatively simple settings. 
For example, in \Figref{fig:width} for \texttt{CNN-VEC}
with NTK parameterization the performance depends non-monotonically on the width, and the optimal width has an intermediate value.\footnote{Similar behavior was observed in~\cite{andreassen2020} for \texttt{CNN-VEC} and in~\cite{aitchison2019bigger} for finite width Bayesian networks.} This nonmonotonicity is distinct from double-descent-like behavior, as all widths correspond to overparameterized models.

\subsection{Diagonal regularization of kernels behaves like early stopping}\label{sec:diag-reg}
\vspace{-0.1cm}
\begin{figure}
\centering
\includegraphics[width=0.245\columnwidth]{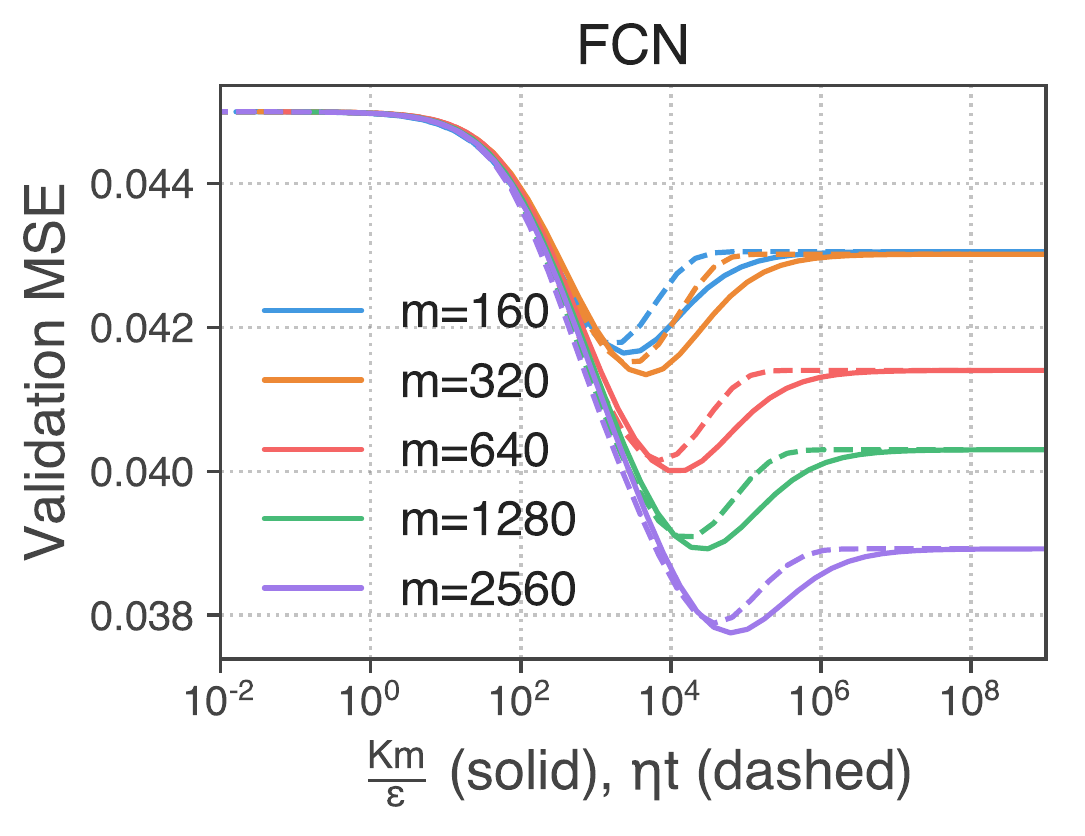}
\includegraphics[width=0.245\columnwidth]{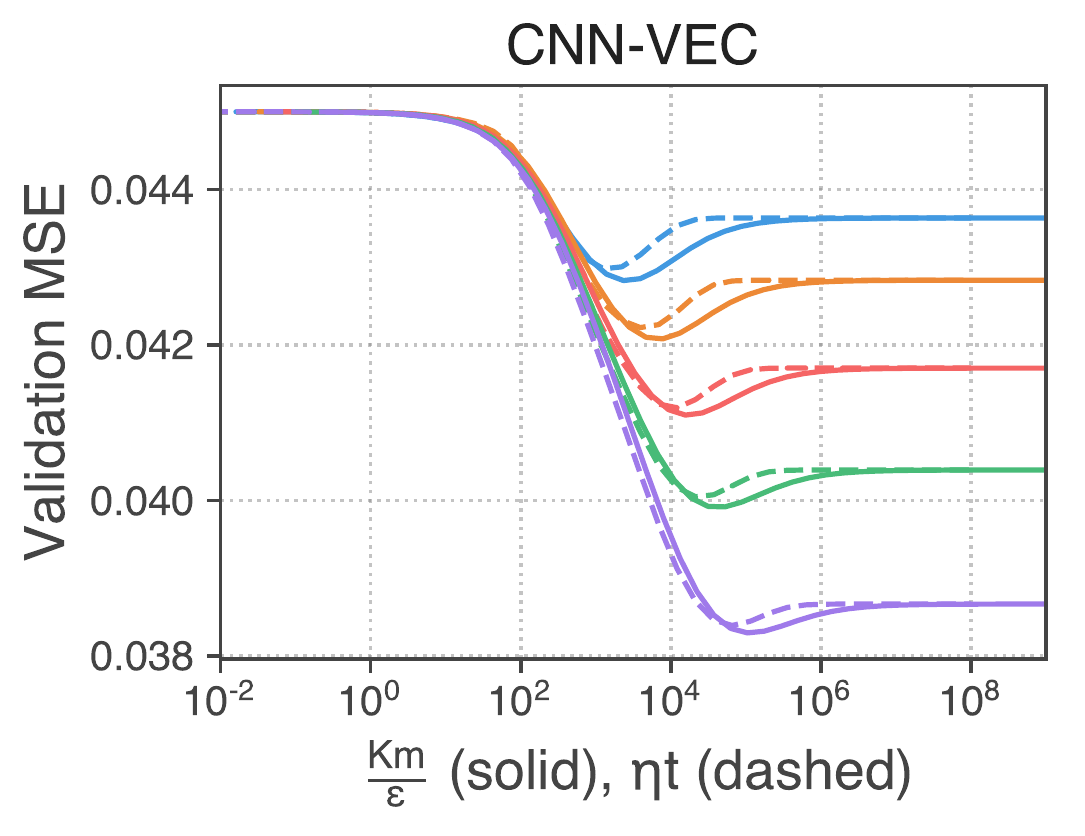}
\includegraphics[width=0.495\columnwidth]{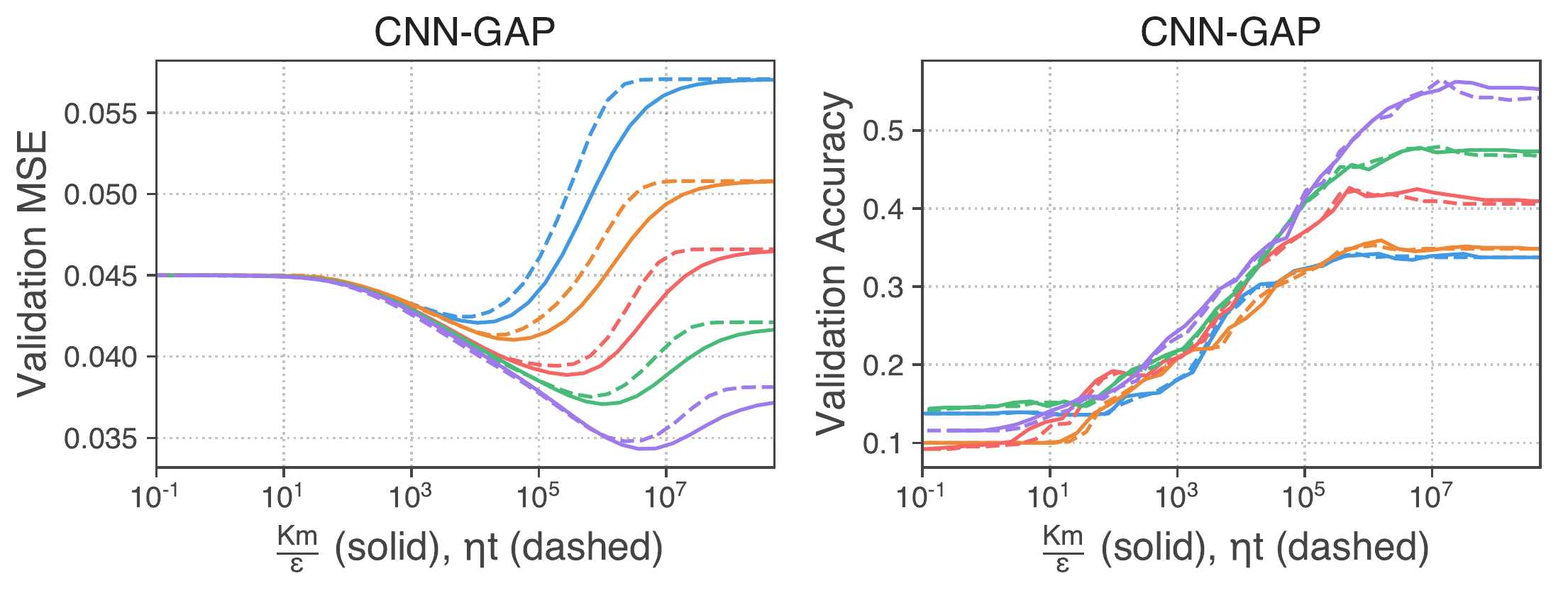}
\caption{
\capsize
\textbf{Diagonal kernel regularization acts similarly to early stopping.}
Solid lines corresponds to NTK inference with varying diagonal regularization $\varepsilon$. Dashed lines correspond to predictions after gradient descent evolution to time $\tau = \eta t$ (with $\eta=\nicefrac{m}{\textrm{tr}({\cal K})}$). 
Line color indicates varying training set size $m$. 
Performing early stopping at time $t$ corresponds closely to regularizing with coefficient $\varepsilon = \nicefrac{K m}{\eta t}$, where $K=10$ denotes number of output classes.
}
\label{fig:diag-reg}
\end{figure}

When performing kernel inference, it is common to add a diagonal regularizer to the training kernel matrix, ${\cal K}_{\textrm{reg}} = {\cal K} + \varepsilon \tfrac{\textrm{tr}({\cal K})}{m} I$. For linear regression, \citet{ali2019continuous} proved that the inverse of a kernel regularizer is related to early stopping time under gradient flow. With kernels, gradient flow dynamics correspond directly to training of a wide neural network \citep{Jacot2018ntk, lee2019wide}. 

We experimentally explore the relationship between early stopping, kernel regularization, and generalization in \Figref{fig:diag-reg}. 
We observe a close relationship between regularization and early stopping, and find that in most cases the best validation performance occurs with early stopping and non-zero $\varepsilon$. 
While \citet{ali2019continuous} do not consider a $\tfrac{\textrm{tr}({\cal K})}{m}$ scaling on the kernel regularizer, we found it useful since experiments become invariant under scale of ${\cal K}$.

\subsection{
Floating point precision determines critical dataset size for failure of kernel methods
}
\label{sec:kernel eigs}

\begin{figure}[t!]
\centering
\includegraphics[width=\linewidth]{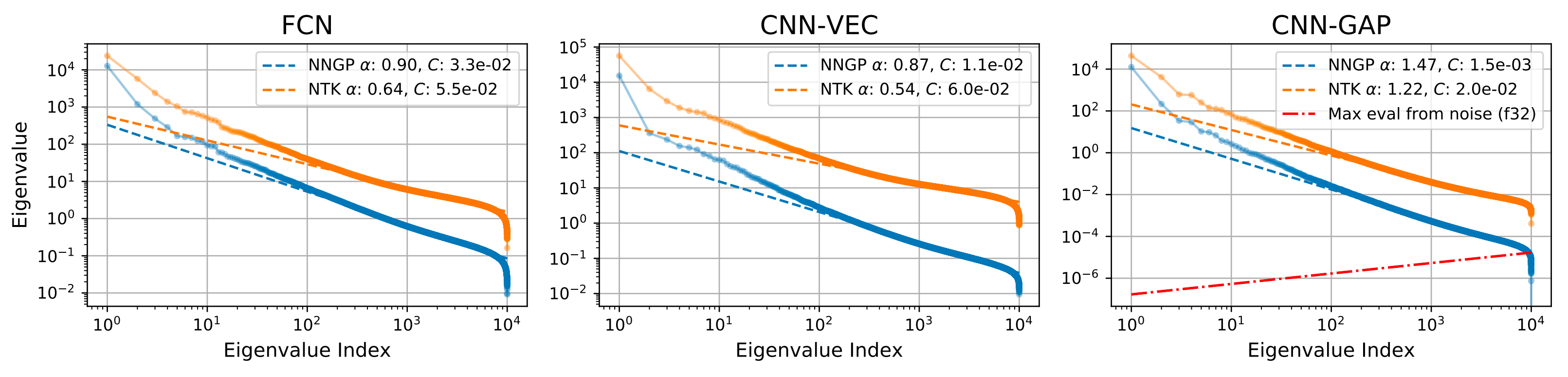}
\caption{
\textbf{Tail eigenvalues of infinite network kernels show power-law decay.} 
The red dashed line shows the predicted scale of noise in the eigenvalues due to floating point precision, for kernel matrices of increasing width. 
Eigenvalues for CNN-GAP architectures decay fast, and may be overwhelmed by \texttt{float32} quantization noise 
for dataset sizes of $O(10^4)$. For \texttt{float64}, quantization noise is not predicted to become significant until a dataset size of $O(10^{10})$ (\Figref{app kernel spectra}).
}
\label{fig:kernel_spectra}
\end{figure}

We observe empirically that kernels become sensitive to \texttt{float32} vs. \texttt{float64} numerical precision at a critical dataset size. For instance, GAP models suffer \texttt{float32} numerical precision errors at a dataset size of $\sim{10}^4$.
 This phenomena can be understood with a simple random noise model (see \sref{app:noise-model} for details). The key insight is that kernels with fast eigenvalue decay suffer from floating point noise. Empirically, the tail eigenvalue of the NNGP/NTK follows a power law (see \Figref{fig:kernel_spectra})
 and measuring their decay trend provides good indication of critical dataset size
\begin{equation}
    m^* \gtrsim
    \left(\nicefrac{C}{\pp{\sqrt{2} \sigma_n}}\right)^{\tfrac{2}{2\alpha - 1}} \quad \textrm{if } \alpha > \tfrac{1}{2}\ \qquad \left(\infty  \quad \textrm{otherwise}\right)\,,
\label{eq:critical-m}
\end{equation}
where $\sigma_n$ is the typical noise scale, e.g. \texttt{float32} epsilon, and the kernel eigenvalue decay  is modeled as $\lambda_i \sim C \, i^{-\alpha}$ as $i$ increases. 
Beyond this critical dataset size, the smallest eigenvalues in the kernel become dominated by floating point noise.

\subsection{Linearized \texttt{CNN-GAP} models perform poorly due to poor conditioning}
\label{sec:cnn-gap-conditioning}

We observe that the linearized \texttt{CNN-GAP} 
converges {\em extremely} slowly on the training set (\Figref{fig:training_curves}), 
leading to poor validation performance (\Figref{fig:validation_curves}). 
Even after training for more than 10M steps with varying L2 regularization strengths and LRs, the best training accuracy was below 90\%, and test accuracy $\sim$70\% -- worse
than both the corresponding infinite and nonlinear finite width networks.

This is caused by 
poor conditioning of pooling networks. \citet{xiao2019disentangling} (Table 1) show that the conditioning at initialization of a \texttt{CNN-GAP} network is worse than that of \texttt{FCN} or \texttt{CNN-VEC} networks by a factor of the number of pixels (1024 for CIFAR-10). This poor conditioning of the kernel eigenspectrum can be seen in \Figref{fig:kernel_spectra}. For linearized networks, in addition to slowing training by a factor of 1024, this leads to numerical instability when using \texttt{float32}.

\subsection{
Regularized ZCA whitening improves accuracy
}\label{sec:zca}

\begin{figure}
\centering
\begin{overpic}[width=\linewidth]{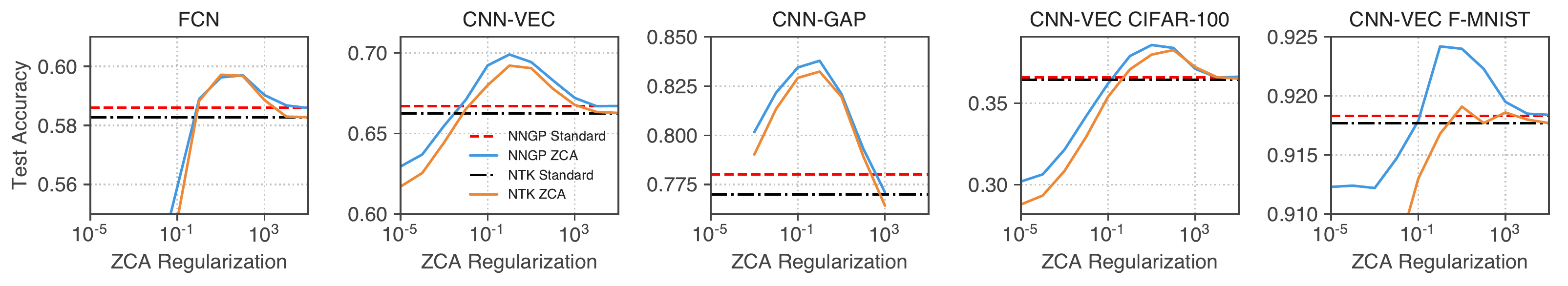}
 \put (0,0) {\textbf{\small(a)}}
\end{overpic}   

\vspace{0.2cm}
\begin{overpic}[width=\linewidth]{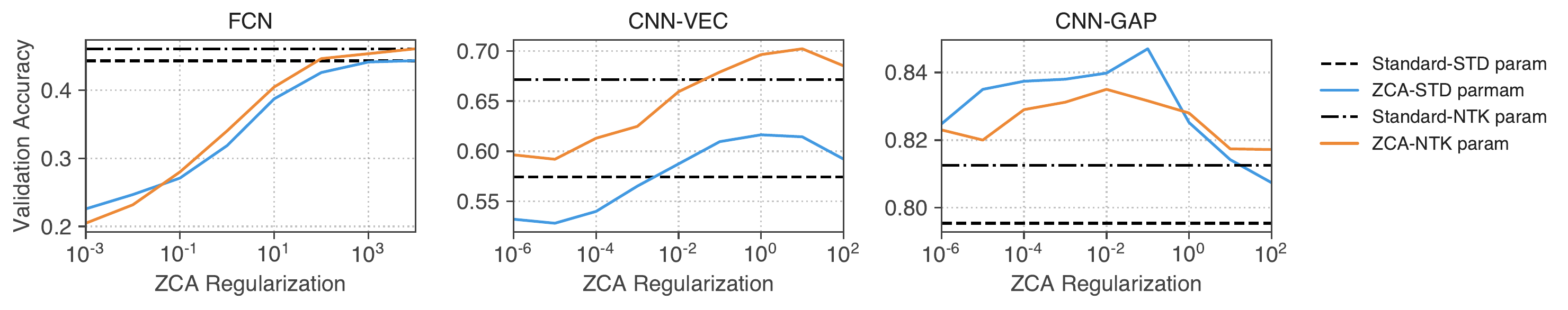}
 \put (0,0) {\textbf{\small(b)}}
\end{overpic}   

\caption{
\capsize
\textbf{Regularized ZCA whitening improves image classification performance for both finite and infinite width networks.} 
All plots show performance as a function of ZCA regularizaiton strength.
(\textbf{a}) ZCA whitening of inputs to kernel methods on CIFAR-10, Fashion-MNIST, and CIFAR-100. (\textbf{b}) ZCA whitening of inputs to finite width networks (training curves in \Figref{fig:app-zca-training}).
}
\label{fig:zca}
\end{figure}

ZCA whitening~\cite{bell1997independent} (see \Figref{fig cifar zca} for an illustration) is a data preprocessing technique that was once common~\cite{goodfellow2013maxout,zagoruyko2016wide}, but has fallen out of favor. However it was recently shown to dramatically improve accuracy in some kernel methods by~\citet{Shankar2020NeuralKW}, in combination with a small regularization parameter in the denominator (see \sref{app ZCA}). 
We investigate the utility of ZCA whitening as a preprocessing step for both finite and infinite width neural networks. 
We observe that while pure ZCA whitening is detrimental for both kernels and finite networks (consistent with predictions in \citep{wadia2020whitening}), with tuning of the regularization parameter it provides performance benefits for both kernel methods and finite network training (\Figref{fig:zca}).

\subsection{Equivariance 
is only beneficial for narrow networks far from the kernel regime
}\label{sec:equivariance}
    
    Due to weight sharing between spatial locations, outputs of a convolutional layer are translation-{\em equivariant} (up to edge effects), i.e. if an input image is translated, the activations are translated in the same spatial direction. However, the vast majority of contemporary \texttt{CNN}s utilize weight sharing in conjunction with pooling layers, making the network outputs approximately translation-\textit{invariant} (\texttt{CNN-GAP}). 
    The impact of equivariance alone (\texttt{CNN-VEC})
    on generalization is not well understood -- 
    it is a property of internal representations only, and does not translate into meaningful statements about the classifier outputs. 
    Moreover, in the infinite-width limit it is guaranteed to have no impact on the outputs \citep{novak2018bayesian, yang2019scaling}. In the finite regime it has been reported both to provide substantial benefits by \citet{lecun1989generalization, novak2018bayesian} and no significant benefits by \citet{bartunov2018assessing}.
    
    We conjecture that equivariance can only be leveraged far from the kernel regime. Indeed, as observed in \Figref{fig:tricks_vs_accuracy} and discussed in \sref{sec:l2_lr}, multiple kernel correspondence-breaking tricks are required for a meaningful boost in performance over NNGP or NTK (which are mathematically guaranteed to not benefit from equivariance), and the boost is largest at a moderate
    width (\Figref{fig:width}). 
    Otherwise, even large ensembles of equivariant models (see \texttt{CNN-VEC LIN} in \Figref{fig:ensemble}) perform comparably to their infinite width, equivariance-agnostic counterparts. Accordingly, prior work that managed to extract benefits from equivariant models \citep{lecun1989generalization, novak2018bayesian} tuned networks far outside the kernel regime (extremely small size and \texttt{+LR+L2+U} respectively). We further confirm this phenomenon in a controlled setting in \Figref{fig:crop_translate}.

    \definecolor{darkred_f}{RGB}{247, 129, 191}
    \definecolor{darkblue_f}{RGB}{55, 126, 184}
    \definecolor{darkorange_f}{RGB}{255, 127, 0}
    \definecolor{darkgreen_f}{RGB}{77, 175, 74}

    \begin{figure}
        \centering
        \includegraphics[width=0.4\textwidth]{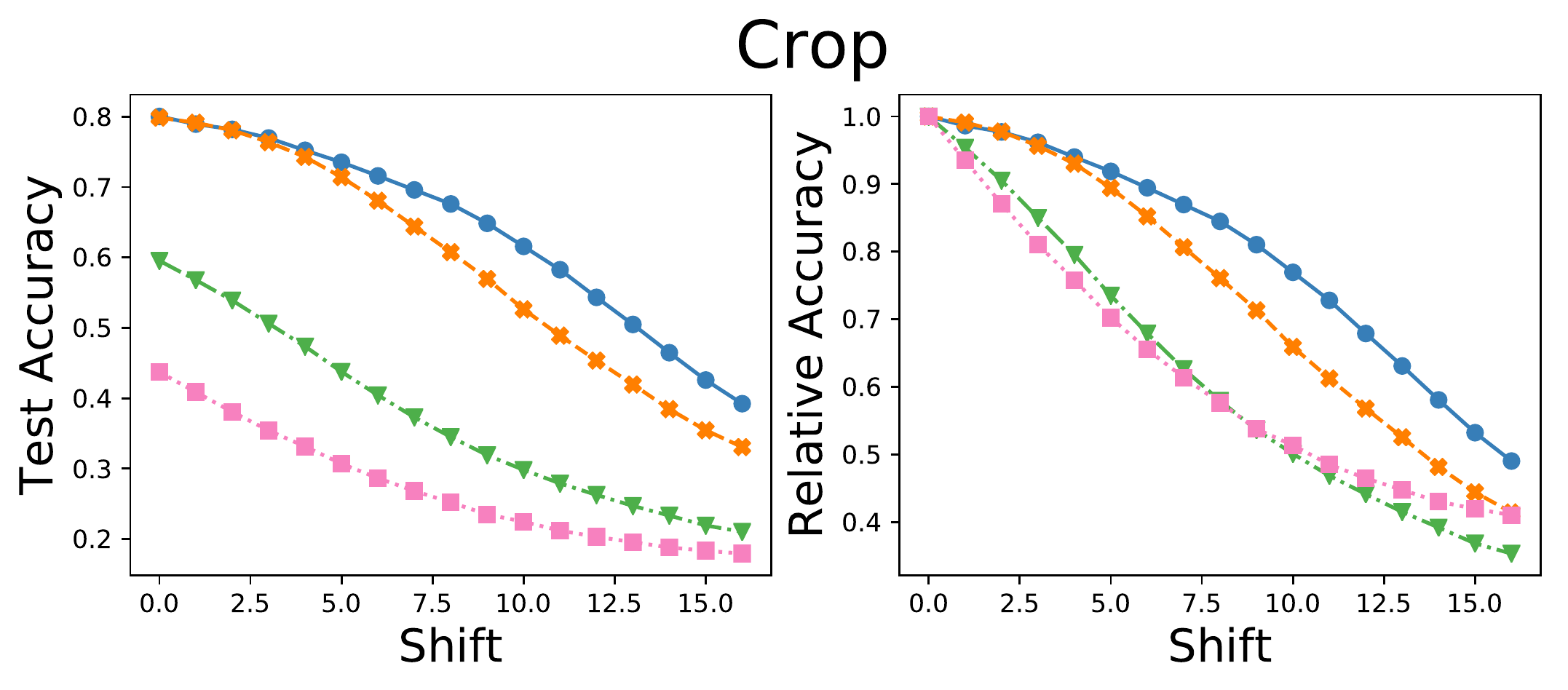}
        \includegraphics[width=0.59\textwidth]{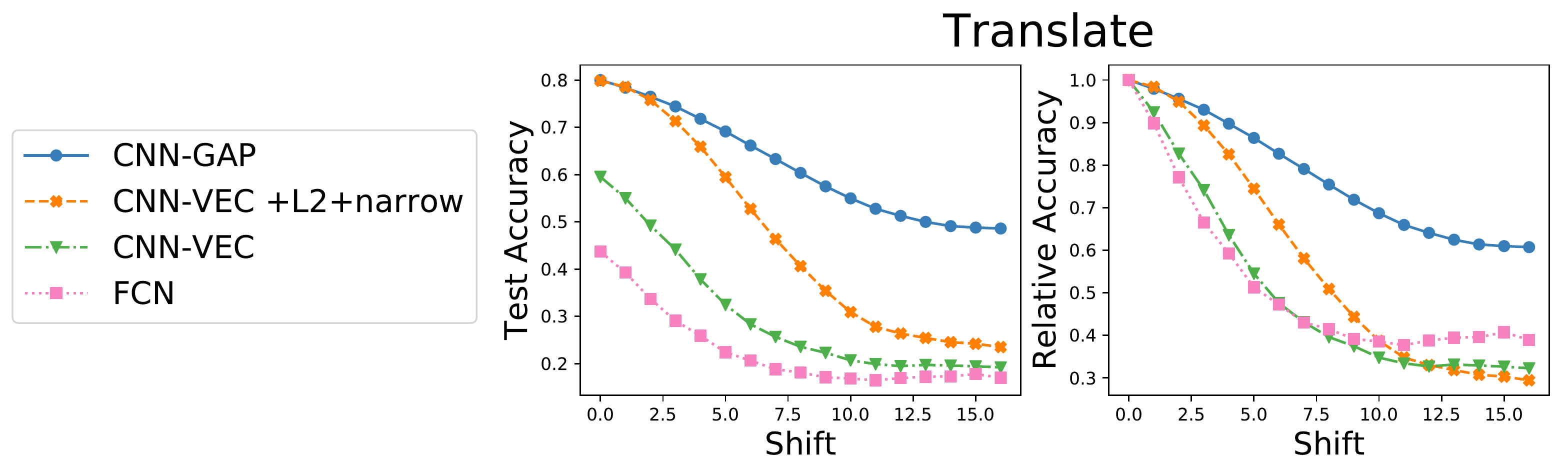}
        \caption{
        \capsize
        \textbf{Equivariance is only leveraged in a \texttt{CNN} model outside of the kernel regime.} 
        If a \texttt{CNN} model is able to utilize equivariance effectively, we expect it to be more robust to crops and translations than an {\color{darkred_f}\texttt{FCN}}. Surprisingly, performance of a {\color{darkgreen_f}wide \texttt{CNN-VEC}} degrades with the  magnitude of the input perturbation as fast as that of an  {\color{darkred_f}\texttt{FCN}}, indicating that equivariance is not exploited.
        In contrast, performance of a {\color{darkorange_f}narrow model with weight decay (\texttt{CNN-VEC+L2+narrow})} falls off much slower.
        {\color{darkblue_f}Translation-invariant \texttt{CNN-GAP}} remains, as expected, the most robust. Details in \sref{sec:equivariance}, \sref{app hyperparameters}. 
        }
        \label{fig:crop_translate}
    \end{figure}

\subsection{Ensembling kernel predictors enables practical data augmentation with NNGP/NTK}\label{sec:data-augmentation}
Finite width neural network often are trained with data augmentation (DA) to improve performance. We observe that the \texttt{FCN} and \texttt{CNN-VEC} architectures (both finite and infinite networks) benefit from DA, and that DA can cause \texttt{CNN-VEC} to become competitive with \texttt{CNN-GAP} (\Tabref{tab:main-table}). While \texttt{CNN-VEC} possess translation equivariance but not invariance (\sref{sec:equivariance}), we believe it can effectively leverage equivariance to learn invariance from data.

For kernels, expanding a dataset with augmentation is computationally challenging, since kernel computation is quadratic in dataset size, and inference is cubic. 
\citet{li2019enhanced, Shankar2020NeuralKW} incorporated flip augmentation by doubling the training set size. 
Extending this strategy to more augmentations such as crop or mixup~\cite{zhang2018mixup}, or to broader augmentations strategies like AutoAugment~\cite{cubuk2019autoaugment} and RandAugment~\cite{cubuk2019randaugment}, becomes rapidly infeasible.

\begin{figure}
\centering
\includegraphics[width=\columnwidth]{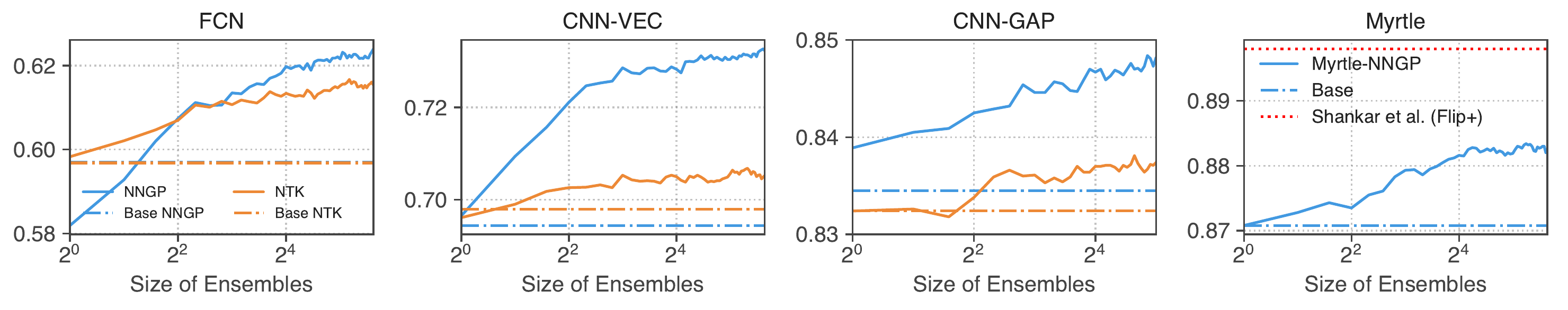}
\caption{
\capsize
\textbf{Ensembling kernel predictors makes predictions from large augmented datasets computationally tractable.} 
We used standard crop by 4 and flip data augmentation (DA) common for training neural networks for CIFAR-10. We observed that DA ensembling improves accuracy and is much more effective for NNGP compared to NTK. In the last panel, we applied data augmentation by ensemble to the Myrtle architecture studied in \citet{Shankar2020NeuralKW}. We observe improvements over our base setting, but do not reach the reported best performance. We believe techniques such as leave-one-out tilt and ZCA augmentation also used in~\cite{Shankar2020NeuralKW} contribute to this difference.}
\label{fig:kerne-da-ens}
\end{figure}

Here we introduce a straightforward method for ensembling kernel predictors to enable more extensive data augmentation. 
More sophisticated approximation approaches such as the Nyström method~\citep{williams2001using} might yield even better performance. 
The strategy involves constructing a set of augmented batches, performing kernel inference for each of them, and then performing ensembling 
of the resulting predictions. This is equivalent to replacing the kernel with a block diagonal approximation, where each block corresponds to one of the batches, and the union of all augmented batches is the full augmented dataset. See \sref{app kernel ensembling} for more details.
This method achieves SOTA for a kernel method corresponding to the infinite width limit of each architecture class we studied (\Figref{fig:kerne-da-ens} and \Tabref{tab:sota-kernel-table}).

\begin{table}
\centering
\caption{
\capsize
\textbf{CIFAR-10 test accuracy for kernels of the  corresponding architecture type}}
\label{tab:sota-kernel-table}
\vspace{0.1cm}
\resizebox{\textwidth}{!}{
\begin{tabular}{@{}llll@{}}
\toprule
Architecture  & Method & 
\begin{tabular}[c]{@{}l@{}}NTK
\end{tabular} & 
\begin{tabular}[c]{@{}l@{}}NNGP
\end{tabular} \\ 
\midrule \midrule
{}{}{\textbf{FC}}             
                                         & \citet{novak2018bayesian}                     & -                 & 59.9          \\
                                         &  ZCA Reg (this work)                      & 59.7             & 59.7          \\
                                         & DA Ensemble (this work)               & \textbf{61.5}     & \textbf{62.4} \\
\midrule
{\textbf{CNN-VEC}}        & \citet{novak2018bayesian}                     & \textbf{-}        & 67.1          \\
                                         & \citet{li2019enhanced}                    & 66.6             & 66.8         \\
                                         & ZCA Reg (this work)                      & 69.8             & 69.4         \\
                                         &  Flip Augmentation, \citet{li2019enhanced}            & 69.9             & 70.5          \\
                                         & DA Ensemble (this work)               & \textbf{70.5}     & \textbf{73.2} \\
\midrule
{\textbf{CNN-GAP}}        & \citet{arora2019on, li2019enhanced}             & 77.6             & 78.5          \\
                                         &  ZCA Reg (this work)                      & 83.2    &  83.5 \\
                                         & Flip Augmentation, \citet{li2019enhanced}           & 79.7             & 80.0          \\
                                         & DA Ensemble (this work)               & \textbf{83.7 (32 ens)}            & \textbf{84.8 (32 ens)}         \\
 \midrule
{\textbf{Myrtle}
\tablefootnote{The normalized Gaussian Myrtle kernel used in~\citet{Shankar2020NeuralKW} does not have a corresponding finite-width neural network, and was additionally tuned on the test set for the case of CIFAR-10.} 
} 
                                         & Myrtle ZCA and Flip Augmentation, \citet{Shankar2020NeuralKW} & -    & \textbf{89.8}               \\
\bottomrule
\end{tabular}
}
\end{table}
\section{Discussion}

We performed an in-depth investigation of the phenomenology of finite and infinite width neural networks 
through a series of controlled interventions. 
We 
quantified
phenomena having to do with 
generalization, architecture dependendence, deviations between infinite and finite networks, numerical stability, data augmentation, data preprocessing, ensembling, network topology, and failure modes of linearization.
We further developed best practices that improve performance for both finite and infinite networks.
We believe our experiments provide firm empirical ground for future studies.

The careful study of other architectural components such as self-attention, normalization, and residual connections would be an interesting extension to this work, especially in light of results such as \citet{Goldblum2020Truth} which empirically observes that the large width behavior of Residual Networks does not conform to the infinite-width limit.
Another interesting future direction would be incorporating systematic finite-width corrections, such as those in~\citet{yaida2019non, Dyer2020Asymptotics, antognini2019finite, huang2019dynamics}.

\section*{Broader Impact}

Developing theoretical understanding of neural networks is crucial both for understanding their biases, and predicting when and how they will fail. 
Understanding biases in models is of critical importance if we hope to prevent them from perpetuating and exaggerating existing racial, gender, and other social biases \citep{hardt2016equality, barocas2016big, doshi2017towards, barocas-hardt-narayanan}. 
Understanding model failure has a direct impact on human safety, as neural networks increasingly do things like drive cars and control the electrical grid~\citep{bojarski2016end, rudin2011machine, ozay2015machine}. 

We believe that wide neural networks are currently the most promising direction for the development of neural network theory. 
We further believe that the experiments we present in this paper will provide empirical underpinnings that allow better theory to be developed. 
We thus believe that this paper will in a small way aid the engineering of safer and more just machine learning models.

\begin{ack}
We thank Yasaman Bahri and Ethan Dyer for discussions and feedback on the project.
We are also grateful to Atish Agarwala and Gamaleldin Elsayed for providing valuable feedbacks on a
draft. 

We acknowledge the Python community~\cite{van1995python} for developing the core set of tools that enabled this work, including NumPy~\cite{numpy}, SciPy~\cite{scipy}, Matplotlib~\cite{matplotlib}, Pandas~\cite{pandas}, Jupyter~\cite{jupyter}, JAX~\cite{jaxrepro}, Neural Tangents~\cite{neuraltangents2020}, Apache Beam~\cite{beam}, Tensorflow datasets~\cite{TFDS} and Google Colaboratory~\cite{colab}.
\end{ack}

\small
\bibliography{references}
\bibliographystyle{unsrtnat}
\normalsize
\onecolumn
\clearpage
\appendix

\begin{center}
\textbf{\large Supplementary Material}
\end{center}

\setcounter{equation}{0}
\setcounter{figure}{0}
\setcounter{table}{0}
\setcounter{page}{1}
\setcounter{section}{0}

\renewcommand{\theequation}{S\arabic{equation}}
\renewcommand{\thefigure}{S\arabic{figure}}
\renewcommand{\thetable}{S\arabic{table}}

\section{Glossary}
We use the following abbreviations in this work:

\begin{itemize}
    \item{\bf L2}: L2 reguarization a.k.a. weight decay;
    \item {\bf LR}: using large learning rate;
    \item {\bf  U}: allowing underfitting; 
    \item {\bf DA}: using data augmentation;
    \item {\bf  C}: centering the network so that the logits are always zero at initialization;
    \item {\bf  Ens}: neural network ensembling logits over multiple initialization;
    \item {\bf  ZCA}: zero-phase component analysis regularization preprocessing;
    \item {\bf FCN}: fully-connected neural network.;
    \item {\bf  CNN-VEC}: convolutional neural network with a vectorized readout layer;
    \item {\bf  CNN-GAP}: convolutional neural network with a global average pooling readout layer;
    \item {\bf  NNGP}: neural network Gaussian process;
    \item {\bf  NTK}: neural tangent kernel.
\end{itemize}

\section{Main table}

\begin{table}[h]
\centering
\caption{\textbf{CIFAR-10 classification accuracy for nonlinear and linearized finite neural networks, as well as for NTK and NNGP kernel methods}.
Starting from \texttt{Base} network of given architecture class described in \sref{sec:experimental_design}, performance change of \textbf{centering} (\texttt{+C}), \textbf{large learning rate} (\texttt{+LR}), allowing \textbf{underfitting} by early stopping (\texttt{+U}), input preprocessing with \textbf{ZCA regularization} (\texttt{+ZCA}), multiple initialization \textbf{ensembling} (\texttt{+Ens}), and some combinations are shown, for {\color{standard_param}\textbf{Standard}} and {\color{ntk_param}\textbf{NTK}} parameterization. See also~\Figref{fig:tricks_vs_accuracy}.
}
\vspace{0.1cm}
\label{tab:main-table}
\resizebox{\columnwidth}{!}{%
\begin{tabular}{@{}lc|ccccccccc|cc|cc@{}}
\toprule
{} &
  Param &
  Base &
  +C &
  +LR &
  +L2 &
  \begin{tabular}[c]{@{}l@{}}+L2 \\ +U\end{tabular}
 &
  \begin{tabular}[c]{@{}c@{}}+L2 \\ +LR\end{tabular} &
  \begin{tabular}[c]{@{}l@{}}+L2 \\+LR\\ +U \end{tabular} &
  +ZCA &
  \begin{tabular}[c]{@{}c@{}} Best \\ w/o DA\end{tabular} &
  +Ens &
  \begin{tabular}[c]{@{}l@{}}+Ens \\+C \end{tabular}
&
 \begin{tabular}[c]{@{}l@{}} +DA\\ +U \end{tabular}&
  \begin{tabular}[c]{@{}l@{}}+DA \\+L2\\ +LR\\ +U\end{tabular} \\
\midrule\midrule
FCN &
  \begin{tabular}[c]{@{}c@{}}STD\\ NTK\end{tabular} &
  \begin{tabular}[c]{@{}c@{}}47.82\\ 46.16\end{tabular} &
  \begin{tabular}[c]{@{}c@{}}53.22\\ 51.74\end{tabular} &
  \begin{tabular}[c]{@{}c@{}}49.07\\ 48.14\end{tabular} &
  \begin{tabular}[c]{@{}c@{}}49.82\\ 54.27\end{tabular} &
  \begin{tabular}[c]{@{}c@{}}49.82\\ 54.27\end{tabular} &
  \begin{tabular}[c]{@{}c@{}}55.32\\ 55.11\end{tabular} &
  \begin{tabular}[c]{@{}c@{}}55.32\\ 55.44\end{tabular} &
  \begin{tabular}[c]{@{}c@{}}44.29\\ 44.86\end{tabular} &
  \begin{tabular}[c]{@{}c@{}}55.90\\ 55.44\end{tabular} &
  \begin{tabular}[c]{@{}c@{}}58.11\\ 58.14\end{tabular} &
  \begin{tabular}[c]{@{}c@{}}58.25\\ 58.31\end{tabular} &
  \begin{tabular}[c]{@{}c@{}} 65.29\\ 61.87\end{tabular} &
  \begin{tabular}[c]{@{}c@{}} 67.43\\ 69.35\end{tabular} \\
 \midrule
CNN-VEC &
  \begin{tabular}[c]{@{}c@{}}STD\\ NTK\end{tabular} &
  \begin{tabular}[c]{@{}c@{}}56.68\\ 60.73\end{tabular} &
  \begin{tabular}[c]{@{}c@{}}60.82\\ 58.09\end{tabular} &
  \begin{tabular}[c]{@{}c@{}}62.16\\ 60.73\end{tabular} &
  \begin{tabular}[c]{@{}c@{}}57.15\\ 61.30\end{tabular} &
  \begin{tabular}[c]{@{}c@{}}67.07\\ 75.85\end{tabular} &
  \begin{tabular}[c]{@{}c@{}}62.16\\ 76.93\end{tabular} &
  \begin{tabular}[c]{@{}c@{}}68.99\\ 77.47\end{tabular} &
  \begin{tabular}[c]{@{}c@{}}57.39\\ 61.35\end{tabular} &
  \begin{tabular}[c]{@{}c@{}}68.99\\ 77.47\end{tabular} &
  \begin{tabular}[c]{@{}c@{}}67.30\\ 71.32\end{tabular} &
  \begin{tabular}[c]{@{}c@{}}65.65\\ 67.23\end{tabular} &
  \begin{tabular}[c]{@{}c@{}}76.73\\ 83.92\end{tabular} &
  \begin{tabular}[c]{@{}c@{}}83.01\\ 85.63\end{tabular} \\
  \midrule
CNN-GAP &
  \begin{tabular}[c]{@{}c@{}}STD\\ NTK\end{tabular} &
  \begin{tabular}[c]{@{}c@{}}80.26\\ 80.61\end{tabular} &
  \begin{tabular}[c]{@{}c@{}}81.25\\ 81.73\end{tabular} &
  \begin{tabular}[c]{@{}c@{}}80.93\\ 82.44\end{tabular} &
  \begin{tabular}[c]{@{}c@{}}81.67\\ 81.17\end{tabular} &
  \begin{tabular}[c]{@{}c@{}}81.10\\ 81.17\end{tabular} &
  \begin{tabular}[c]{@{}c@{}}83.69\\ 82.44\end{tabular} &
  \begin{tabular}[c]{@{}c@{}}83.01\\ 82.43\end{tabular} &
  \begin{tabular}[c]{@{}c@{}}84.90\\ 83.75\end{tabular} &
  \begin{tabular}[c]{@{}c@{}}84.22\\ 83.92\end{tabular} &
  \begin{tabular}[c]{@{}c@{}}84.15\\ 85.22\end{tabular} &
  \begin{tabular}[c]{@{}c@{}}84.62\\ 85.75\end{tabular} &
  \begin{tabular}[c]{@{}c@{}}84.36\\ 84.07\end{tabular} &
  \begin{tabular}[c]{@{}c@{}}86.45\\ 86.68\end{tabular}
\end{tabular}%
}
\\
\vspace{0.2cm}
\resizebox{\columnwidth}{!}{%
\begin{tabular}{@{}lc|cccccc||ccc|ccc@{}}
\toprule
 &
  Param &
  Lin Base &
  +C &
  +L2 &
  \begin{tabular}[c]{@{}l@{}}+L2 \\+U \end{tabular} &
  +Ens &
  \begin{tabular}[c]{@{}l@{}}+Ens \\+C \end{tabular}&
  NTK & +ZCA & \begin{tabular}[c]{@{}c@{}}+DA \\ +ZCA\end{tabular}&
  NNGP &+ZCA &
\begin{tabular}[c]{@{}c@{}}+DA \\ +ZCA\end{tabular}\\
  \midrule\midrule
FCN &
  \begin{tabular}[c]{@{}c@{}}STD\\ NTK\end{tabular} &
  \begin{tabular}[c]{@{}c@{}}43.09\\ 48.61\end{tabular} &
  \begin{tabular}[c]{@{}c@{}}51.48\\ 52.12\end{tabular} &
  \begin{tabular}[c]{@{}c@{}}44.16\\ 51.77\end{tabular} &
  \begin{tabular}[c]{@{}c@{}}50.77\\ 51.77\end{tabular} &
  \begin{tabular}[c]{@{}c@{}}57.85\\ 58.04\end{tabular} &
  \begin{tabular}[c]{@{}c@{}}57.99\\ 58.16\end{tabular} &
  \begin{tabular}[c]{@{}c@{}}58.05\\ 58.28\end{tabular} &
  \begin{tabular}[c]{@{}c@{}}59.65\\ 59.68\end{tabular}  &
  \begin{tabular}[c]{@{}c@{}}-\\ 61.54\end{tabular} &
  58.61 &
  59.70 &
  62.40 \\
  \midrule
CNN-VEC &
  \begin{tabular}[c]{@{}c@{}}STD\\ NTK\end{tabular} &
  \begin{tabular}[c]{@{}c@{}}52.43\\ 55.88\end{tabular} &
  \begin{tabular}[c]{@{}c@{}}60.61\\ 58.94\end{tabular} &
  \begin{tabular}[c]{@{}c@{}}58.41\\ 58.52\end{tabular} &
  \begin{tabular}[c]{@{}c@{}}58.41\\ 58.50\end{tabular} &
  \begin{tabular}[c]{@{}c@{}}64.58\\ 65.45\end{tabular} &
  \begin{tabular}[c]{@{}c@{}}64.67\\ 65.54\end{tabular} &
  {\begin{tabular}[c]{@{}c@{}}66.64\\ 66.78\end{tabular}} &
  \begin{tabular}[c]{@{}c@{}}69.65\\ 69.79\end{tabular}&
  \begin{tabular}[c]{@{}c@{}}-\\ 70.52\end{tabular} &
  66.69 &
  69.44 &
  73.23 \\
  \midrule
CNN-GAP &
  \begin{tabular}[c]{@{}c@{}}STD\\ NTK\end{tabular} &
  \multicolumn{6}{c||}{\begin{tabular}[c]{@{}c@{}} \textgreater 70.00* (Train accuracy 86.22 after 14M steps)\\ \textgreater 68.59* (Train accuracy 79.90 after 14M steps)\end{tabular}} &
  \begin{tabular}[c]{@{}c@{}}76.97\\ 77.00\end{tabular} &
  \begin{tabular}[c]{@{}c@{}}83.24\\ 83.24\end{tabular} &
   \begin{tabular}[c]{@{}c@{}}-\\ 83.74\end{tabular}
   &
  78.0 &
  83.45 & 84.82
\end{tabular}%
}
\end{table}

\section{Experimental details}

For all experiments, we use Neural Tangents (NT) library~\cite{neuraltangents2020} built on top of JAX~\cite{jaxrepro}. First we describe experimental settings that is mostly common and then describe specific details and hyperparameters for each experiments.

\textbf{Finite width neural networks}
We train finite width networks with Mean Squared Error (MSE) loss 
$${\cal L } = \frac{1}{2 |{\cal D}| K} \sum_{(x_i, y_i)\in {\cal D}} \|f(x_i) - y_i\|^2\,,$$
where $K$ is the number of classes and $\|\cdot \|$ is the $L^2$ norm in $\mathbb R^{K}$. For the experiments with \texttt{+L2}, we add L2 regularization to the loss 
\begin{equation}\label{eq:l2-reg}
    {R}_{\text{L2}} = \frac{\lambda}{2} \sum_l  \norm{W^l}^2\,,
\end{equation}
and tune $\lambda$ using grid-search optimizing for the validation accuracy.

We optimize the loss using mini-batch SGD with constant learning rate. We use batch-size of $100$ for \texttt{FCN} and $40$ for both \texttt{CNN-VEC} and \texttt{CNN-GAP} (see \sref{app:batch-size} for further details on this choice). 
Learning rate is parameterized with learning rate factor $c$ with respect to the critical learning rate
\begin{equation}
    \eta = c\,  \eta_\text{critical}\,.
\end{equation}
In practice, we compute empirical NTK $\hat \Theta (x, x') = \sum_j \partial_j f(x) \partial_j f(x')$ on 16 random points in the training set to estimate $\eta_\text{critical}$~\cite{lee2019wide} by maximum eigenvalue of $\hat \Theta (x, x)$. This is readily available in NT library~\cite{neuraltangents2020} using \texttt{nt.monte\_carlo\_kernel\_fn} and \texttt{nt.predict.max\_learning\_rate}. 
Base case considered without large learning rate indicates $c \leq 1$, and large learning rate (\texttt{+LR}) runs are allowing $c > 1$. Note that for linearized networks $\eta_\text{critical}$ is strict upper-bound for the learning rates and no $c >1$ is allowed~\cite{lee2019wide, yang2019fine, lewkowycz2020large}.

Training steps are chosen to be large enough, such that learning rate factor $c \leq 1$ can reach above $99\%$ accuracy on $5k$ random subset of training data for 5 logarithmic spaced measurements. For different learning rates, physical time $t=\eta \times \text{(\# of steps)}$ roughly determines learning dynamics and small learning rate trials need larger number of steps. 
Achieving termination criteria was possible for all of the trials except for linearized \texttt{CNN-GAP} and data augmented training of \texttt{FCN}, \texttt{CNN-VEC}. In these cases, we report best achieved performance without fitting the training set.

\textbf{NNGP / NTK} For inference, except for data augmentation ensembles for which default zero regularization was chosen, we grid search over diagonal regularization in the range \texttt{numpy.logspace(-7, 2, 14)} and $0$. Diagonal regularization is parameterized as
$${\cal K}_{\textrm{reg}} = {\cal K} + \varepsilon \tfrac{\textrm{tr}({\cal K})}{m} I$$
where ${\cal K}$ is either NNGP or NTK for the training set. We work with this parameterization since $\varepsilon$ is invariant to scale of ${\cal K}$.

\textbf{Dataset}
For all our experiments (unless specified) we use train/valid/test split of 45k/5k/10k  for CIFAR-10/100 and 50k/10k/10k for Fashion-MNIST. For all our experiments, inputs are standardized with per channel mean and standard deviation. ZCA regularized whitening is applied as described in \sref{app ZCA}. 
Output is encoded as mean subtracted one-hot-encoding for the MSE loss, e.g. for a label in class $c$, $-0.1 \cdot \bf{1} + e_c$. For the softmax-cross-entropy loss in~\sref{app:xent-vs-mse}, we use standard one-hot-encoded output.

For data augmentation, we use widely-used augmentation for CIFAR-10; horizontal flips with 50\% probability and random crops by 4-pixels with zero-padding.

\textbf{Details of architecture choice:}
We only consider ReLU activation (with the exception of Myrtle-kernel which use scaled Gaussian activation~\cite{Shankar2020NeuralKW}) and choose critical initialization weight variance of $\sigma_w^2=2$ with small bias variance $\sigma_b^2=0.01$. 
For convolution layers, we exclusively consider $3 \times 3$ filters with stride $1$ and \texttt{SAME} (zero) padding so that image size does not change under convolution operation.

\subsection{Hyperparameter configurations for all experiments}
\label{app hyperparameters}

We used grid-search for tuning hyperparameters and use accuracy on validation set for deciding on  hyperparameter configuration or measurement steps (for underfitting / early stopping). All reported numbers unless specified is test set performance.

\textbf{\Figref{fig:tricks_vs_accuracy}, Table~\ref{tab:main-table}}: We grid-search over L2 regularization strength $\lambda \in \{0\} \cup \{10^{-k} | k \text{ from -9 to -3}\}$ and learning rate factor $c \in \{2^k | k\text{ from -2  to  5}\}$. For linearized networks same search space is used except that $c>1$ configuration is infeasible and training diverges. For non-linear, centered runs $c \in \{2^k | k\text{ from 0  to  4}\}$ is used. Network ensembles uses base configuration with $\lambda=0$, $c=1$ with 64 different initialization seed. Kernel ensemble is over 50 predictors for \texttt{FCN} and \texttt{CNN-VEC} and 32 predictors for \texttt{CNN-GAP}. Finite networks trained with data-augmentation has different learning rate factor range of $c \in \{1, 4, 8\}$.

\textbf{\Figref{fig:nngp-vs-ntk}}: Each datapoint corresponds to either standard preprocessed or ZCA regularization preprocessed (as described in~\sref{sec:zca}) with regularization strength was varied in $\{10^{-k}| k \in [-6, -5, ..., 4, 5]\}$ for \texttt{FCN} and \texttt{CNN-VEC}, $\{10^{-k}| k \in [-3, -2, ..., 2, 3]\}$ for \texttt{CNN-GAP}.

\begin{figure}[h]
\centering
\includegraphics[width=\columnwidth]{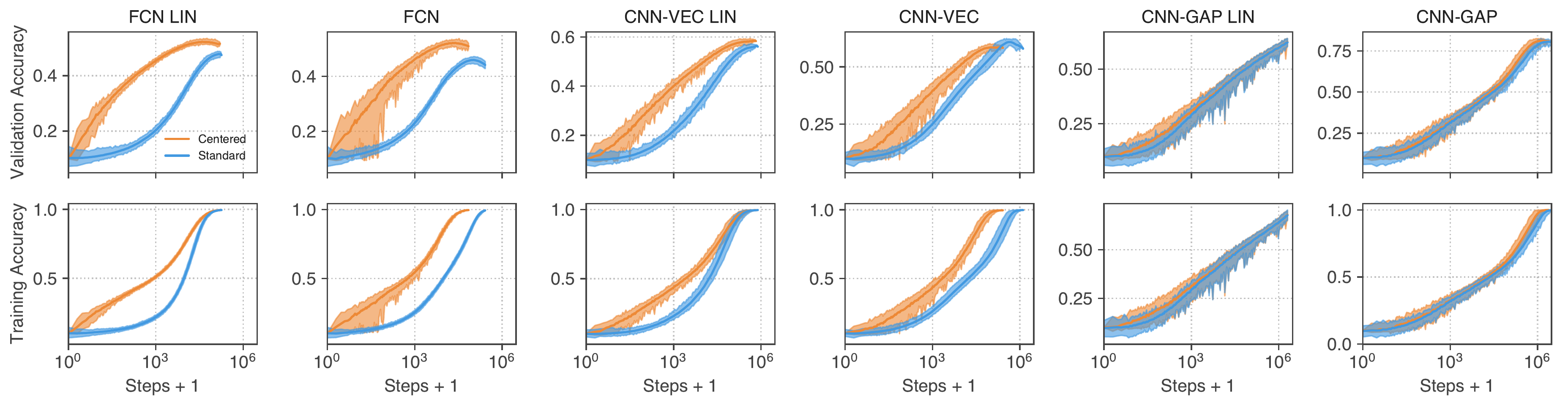}
\caption{\textbf{Centering can accelerate training}. Validation (top) and training (bottom) accuracy throughout training for several finite width architectures. See also \sref{sec:ensemble_of_networks} and \Figref{fig:validation_curves}.
}
\label{fig:training_curves}
\end{figure}

\begin{figure}[h]
\centering
\includegraphics[width=\columnwidth]{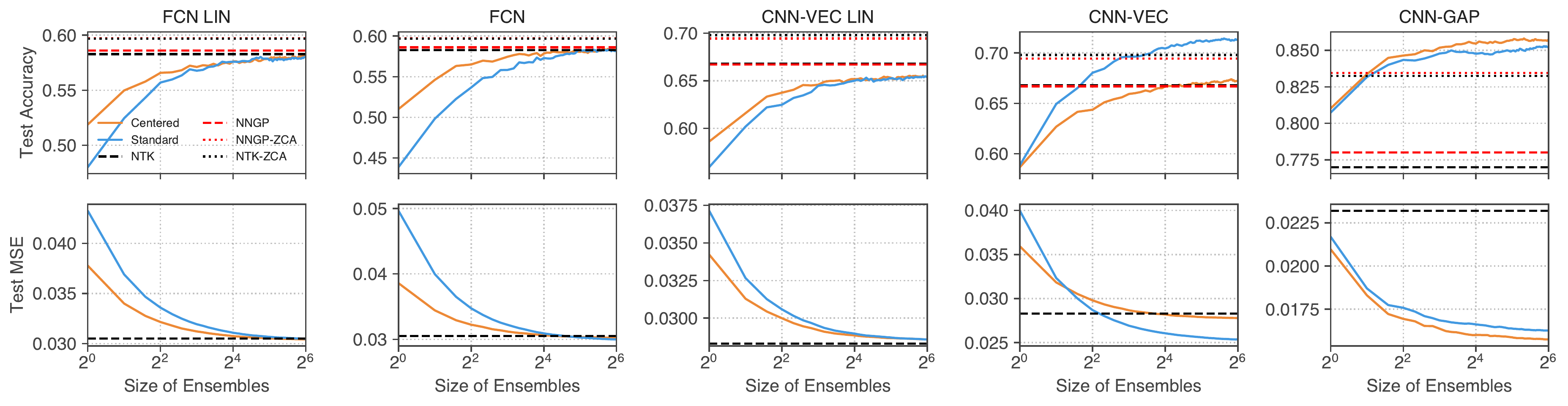}
\caption{\textbf{Ensembling base networks causes them to match kernel performance, or exceed it for nonlinear CNNs.} See also \sref{sec:ensemble_of_networks} and \Figref{fig:ensemble}.
}
\label{fig app ensemble}
\end{figure}

\textbf{\Figref{fig:validation_curves}, \Figref{fig:ensemble}, \Figref{fig:training_curves}, \Figref{fig app ensemble}}: Learning rate factors are $c=1$ for non-linear networks and $c=0.5$ for linearized networks. While we show NTK parameterized runs, we also observe similar trends for STD parameterized networks. Shaded regions show range of minimum and maximum performance across 64 different seeds. Solid line indicates the mean performance. 

\begin{figure}
\centering
\includegraphics[width=\columnwidth]{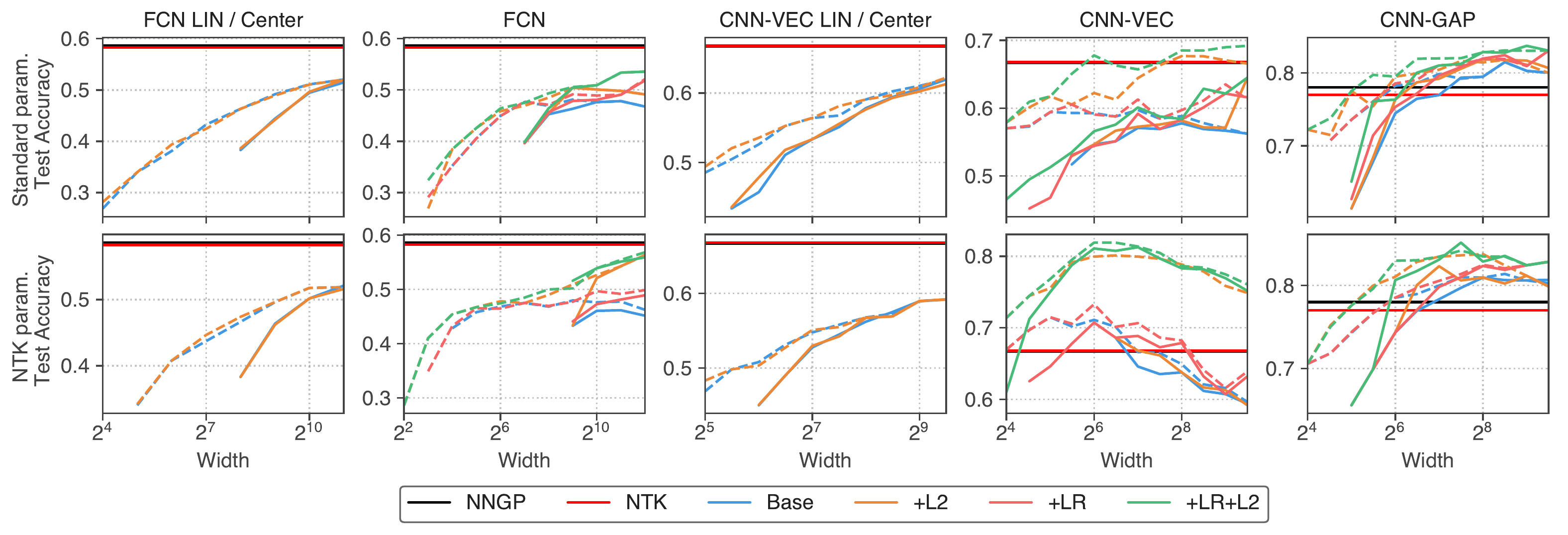}
\caption{
\textbf{Finite width networks generally perform better with increasing width, but \texttt{CNN-VEC} shows surprising non-monotonic behavior.} See also~\sref{sec:perf_vs_width} and \Figref{fig:width}
 {\bf L2}: non-zero weight decay allowed during training {\bf LR}: large learning rate allowed. Dashed lines are allowing underfitting (\textbf{U}).}
\label{fig:width-combined}
\end{figure}

\textbf{\Figref{fig reg compare}}
While \texttt{FCN} is the base configuration, \texttt{CNN-VEC} is a narrow network with 64 channels per layer since moderate width benefits from L2 more for the NTK parameterization~\Figref{fig:width-combined}. For \texttt{CNN-GAP} 128 channel networks is used. All networks with different L2 strategy are trained with \texttt{+LR} ($c>1$).

\begin{figure}[h]
\centering
\begin{overpic}[width=\linewidth]{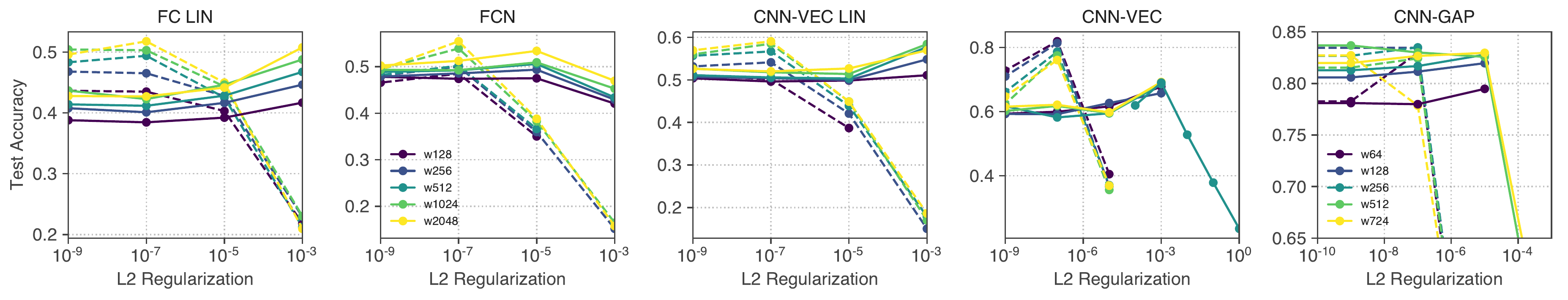}
\end{overpic}    
\\
\caption{
\textbf {Performance of nonlinear and linearized networks as a function of L2 regularization for a variety of widths.} Dashed lines are NTK parameterized networks while solid lines are networks with standard parameterization. We omit linearized \texttt{CNN-GAP} plots as they did not converge even with extensive compute budget.
L2 regularization is more helpful in networks with an NTK parameterization than a standard parameterization 
\label{fig:reg-compare-sm}
}
\end{figure}

\textbf{\Figref{fig:width}, \Figref{fig:reg-compare-sm},  \Figref{fig:width-combined}}: 
$\lambda \in \{0, 10^{-9}, 10^{-7}, 10^{-5}, 10^{-3}\}$ and $c \in \{2^k | k \text{ from} -2 \text{ to } 5\}$.

\textbf{\Figref{fig:diag-reg}}: We use 640 subset of validation set for evaluation. \texttt{CNN-GAP} is a variation of the base model with 3 convolution layers with $\sigma_b^2 = 0.1$ while \texttt{FCN} and \texttt{CNN-VEC} is the base model.
Training evolution is computed using analytic time-evolution described in~\citet{lee2019wide} and implemented in NT library via \texttt{nt.predict.gradient\_descent\_mse} with 0 diagonal regularization.

\textbf{\Figref{fig:zca}}: Kernel experiments details are same as in \Figref{fig:nngp-vs-ntk}. Finite networks are base configuration with $c=1$ and $\lambda=0$. 

\textbf{\Figref{fig:crop_translate}}: Evaluated networks uses NTK parameterization with $c=1$. {\color{darkorange_f}\texttt{CNN-VEC+L2+narrow}} uses 128 channels instead of 512 of the base {\color{darkgreen_f}\texttt{CNN-VEC}} and {\color{darkblue_f}\texttt{CNN-GAP}} networks, and trained with L2 regularization strength $\lambda=10^{-7}$. \emph{Crop} transformation uses zero-padding while \emph{Translate} transformation uses circular boundary condition after shifting images. Each transformation is applied to the test set inputs where shift direction is chosen randomly. Each points correspond to average accuracy over 20 random seeds. {\color{darkred_f}\texttt{FCN}} had 2048 hidden units.

\textbf{\Figref{fig:kerne-da-ens}, Table~\ref{tab:sota-kernel-table}}: For all data augmentation ensembles, first instance is taken to be from non-augmented training set. Further details on kernel ensemble is described in~\sref{app kernel ensembling}. For all kernels, inputs are preprocessed with optimal ZCA regularization observed in~\Figref{fig:zca} (10 for \texttt{FCN}, 1 for \texttt{CNN-VEC}, \texttt{CNN-GAP} and \texttt{Myrtle}.). We ensemble over 50 different augmented draws for \texttt{FCN} and \texttt{CNN-VEC}, whereas for \texttt{CNN-GAP}, we ensemble over 32 draws of augmented training set.

\textbf{\Figref{fig:xent-vs-mse}, Table~\ref{tab:xent-vs-mse}}:
Details for MSE trials are same as ~\Figref{fig:tricks_vs_accuracy} and Table~\ref{tab:main-table}. Trials with softmax-cross-entropy loss was tuned with same hyperparameter range as MSE except that learning rate factor range was $c\in \{1, 4, 8\}$.

\textbf{\Figref{fig:bs}}: We present result with NTK parameterized networks with $\lambda=0$. \texttt{FCN} network is width 1024 with $\eta=10.0$ for MSE loss and $\eta=2.0$ for softmax-cross-entropy loss. \texttt{CNN-GAP} uses 256 channels with $\eta=5.0$ for MSE loss and $\eta=0.2$ for softmax-cross-entropy loss. Random seed was fixed to be the same across all runs for comparison.

\begin{figure}[h]
\centering
\begin{overpic}[width=.6\columnwidth]{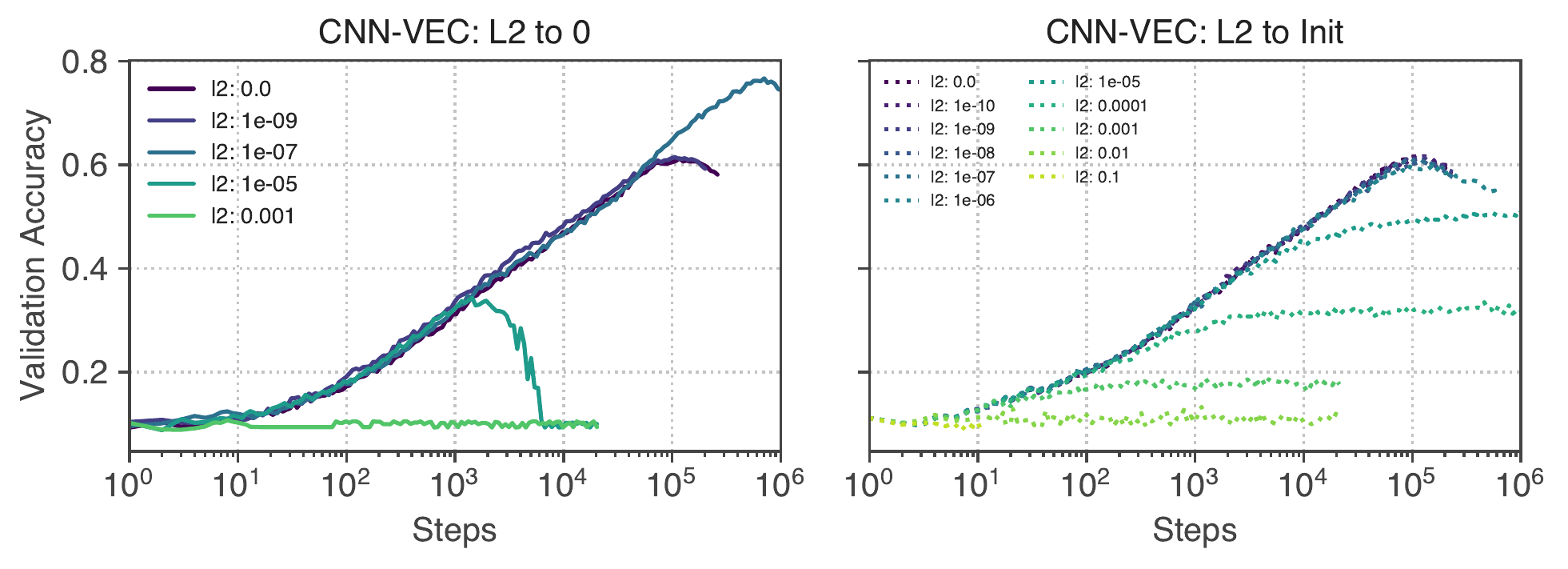}
 \put (0,0) {\textbf{\small(a)}}
\end{overpic} 
\begin{overpic}[width=.32\columnwidth]{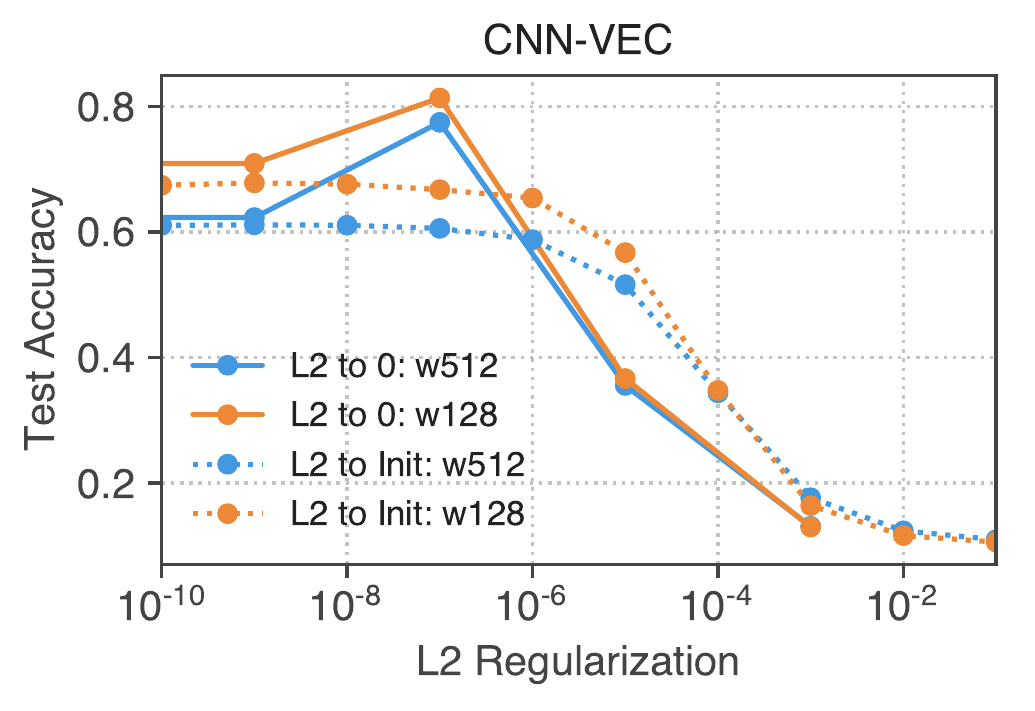}
 \put (0,0) {\textbf{\small(b)}}
\end{overpic}    
\caption{\textbf{L2 regularization to initial weights does not provide performance benefit.} {\bf (a)} Comparing training curves of L2 regularization to either 0 or initial weights. {\bf (b)} Peak performance of after L2 regularization to either 0 or initial weights. Increasing L2 regularization to initial weights do not provide performance benefits, instead performance remains flat until model's capacity deteriorates. 
}
\label{fig:l2-init}
\end{figure}

\textbf{\Figref{fig:l2-init}}: NTK pamareterization with $c=4$ was used for both L2 to zero or initialization. Random seed was fixed to be the same across all runs for comparison.

\section{Noise model}
\label{app:noise-model}
In this section, we provide details on noise model discussed in~\sref{sec:kernel eigs}. Consider a random $m \times m$ Hermitian matrix $N$ with entries order of $\sigma_n$ which is considered as noise perturbation to the kernel matrix
\begin{equation}
    \tilde K = K + N\,.
\end{equation}
Eigenvalues of this random matrix $N$ follow Wigner's semi-circle law, and the smallest eigenvalue is given by $\lambda_{\min}(N) \approx - \sqrt{2m} \sigma_n$. When the smallest eigenvalue of $K$ is smaller (in order) than $|\lambda_{\min}(N)|$, one needs to add diagonal regularizer larger than the order of $|\lambda_{\min}(N)|$ to ensure positive definiteness. For estimates, let us use machine precision\footnote{\texttt{np.finfo(np.float32).eps}, \texttt{np.finfo(np.float64).eps}} $\epsilon_{32} \approx 10^{-7}$  and $\epsilon_{64} \approx 2 \times 10^{-16}$ which we use as proxy values for $\sigma_n$. 
Note that noise scale is relative to elements in $K$ which is assume to be $O(1)$. Naively scaling $K$ by multiplicative constant will also scale $\sigma_n$.

Empirically one can model tail $i^{\textrm{th}}$ eigenvalues of infinite width kernel matrix of size $m \times m$ as
\begin{equation}
    \lambda_i \approx C \frac{ m} {i^{\alpha}} \,.
\end{equation}
Note that we are considering $O(1)$ entries for $K$ and typical eigenvalues scale linearly with dataset size $m$. For a given dataset size, the power law observed is $\alpha$ and $C$ is dataset-size independent constant. Thus the smallest eigenvalue is order $\lambda_\text{min}(K) \sim C m^{1- \alpha}$. 

In the noise model, we can apply Weyl's inequality which says
\begin{equation}
    \lambda_\text{min}(K) - \sqrt{2 m} \sigma_n \leq \lambda_\text{min} (\tilde K ) \leq \lambda_\text{min}(K)  + \sqrt{2 m} \sigma_n \,.
\end{equation}

Consider the worst-case where negative eigenvalue noise affecting the kernel's smallest eigenvalue. In that case perturbed matrices minimum eigenvalue could become negative, breaking positive semi-definiteness(PSD) of the kernel.  

This model allows to predict critical dataset size ($m^*$) over which PSD can be broken under specified noise scale and kernel eigenvalue decay. With condition that perturbed smallest eigenvalue becomes negative 
\begin{equation}
    C m^{1-\alpha} \lesssim \sqrt{2 m}\sigma_n\,,
\end{equation}
we obtain
\begin{equation}
    m^* \gtrsim
    \begin{cases}
    \left(\frac{C}{\sqrt{2} \sigma_n}\right)^{\tfrac{2}{2\alpha - 1}} & \textrm{if } \alpha > \tfrac{1}{2}\\
    \infty & \textrm{else}
    \end{cases}
\end{equation}

When PSD is broken, one way to preserve PSD is to add diagonal regularizer (\sref{sec:diag-reg}). 
For CIFAR-10 with $m=50k$, typical negative eigenvalue from \texttt{float32} noise is around $4 \times 10^{-5}$ and $7 \times 10^{-14}$  with \texttt{float64} noise scale, considering $\sqrt{2 m} \sigma_n$.  Note that \citet{arora2019on} regularized kernel with regularization strength $5 \times 10^{-5}$ which is on par with typical negative eigenvalue introduced due to \texttt{float32} noise. Of course, this only applies if kernel eigenvalue decay is sufficiently fast that full dataset size is above $m^*$.

We observe that \texttt{FCN} and \texttt{CNN-VEC} kernels with small $\alpha$ would not suffer from increasing dataset-size under \texttt{float32} precision. On the other-hand, worse conditioning of \texttt{CNN-GAP} not only affects the training time (\sref{sec:cnn-gap-conditioning}) but also required precision. One could add sufficiently large diagonal regularization to mitigate effect from the noise at the expense of losing information and generalization strength included in eigen-directions with small eigenvalues.

\begin{figure}[t!]
\centering

\begin{overpic}[width=0.86\linewidth]{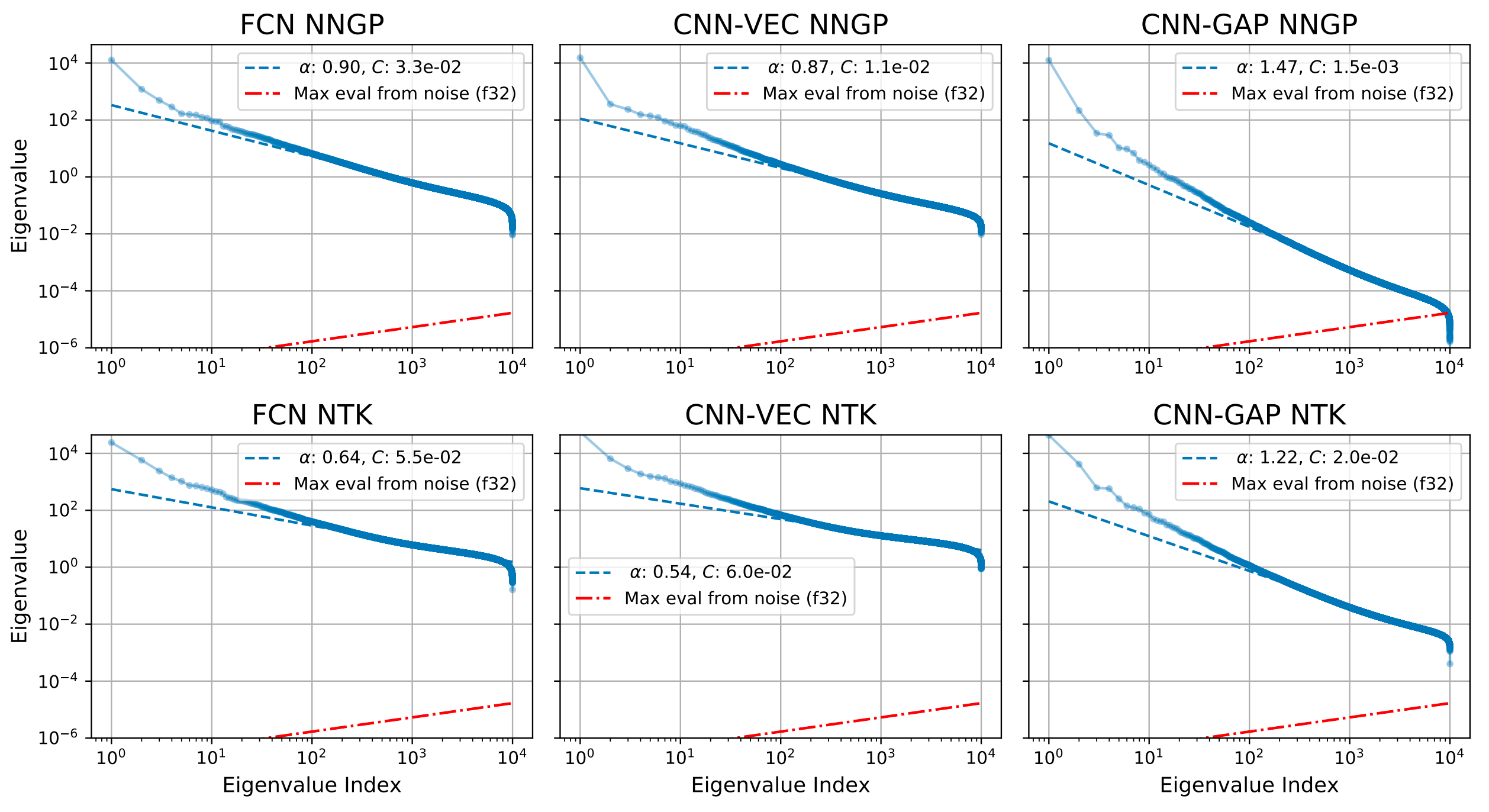}
 \put (0,0) {\textbf{\small(a)}}
 \end{overpic}\\
\vspace{0.3cm}
   \begin{overpic}[width=0.43\linewidth]{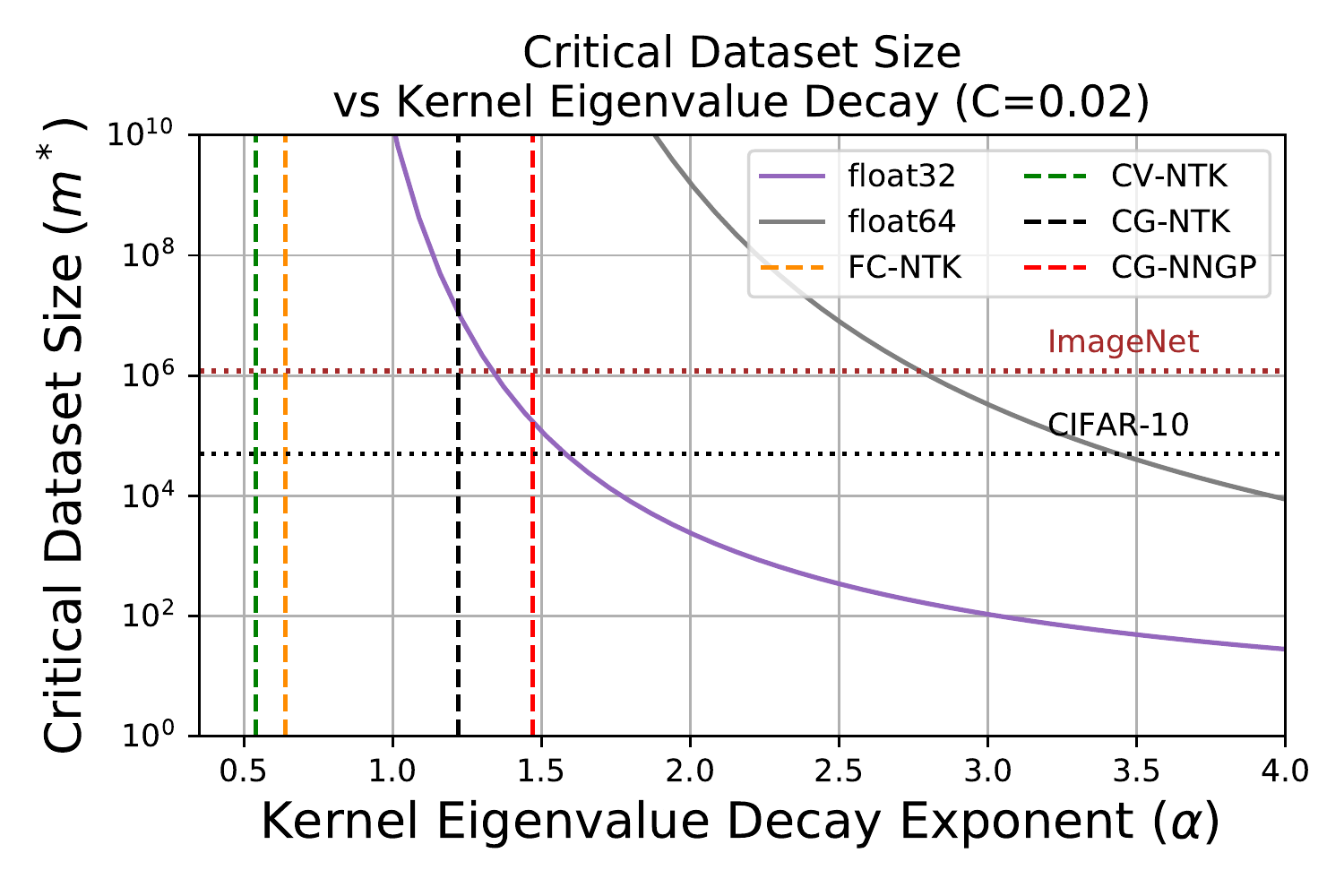}
 \put (0,0) {\textbf{\small(b)}}
 \end{overpic}
 \begin{overpic}[width=0.43\linewidth]{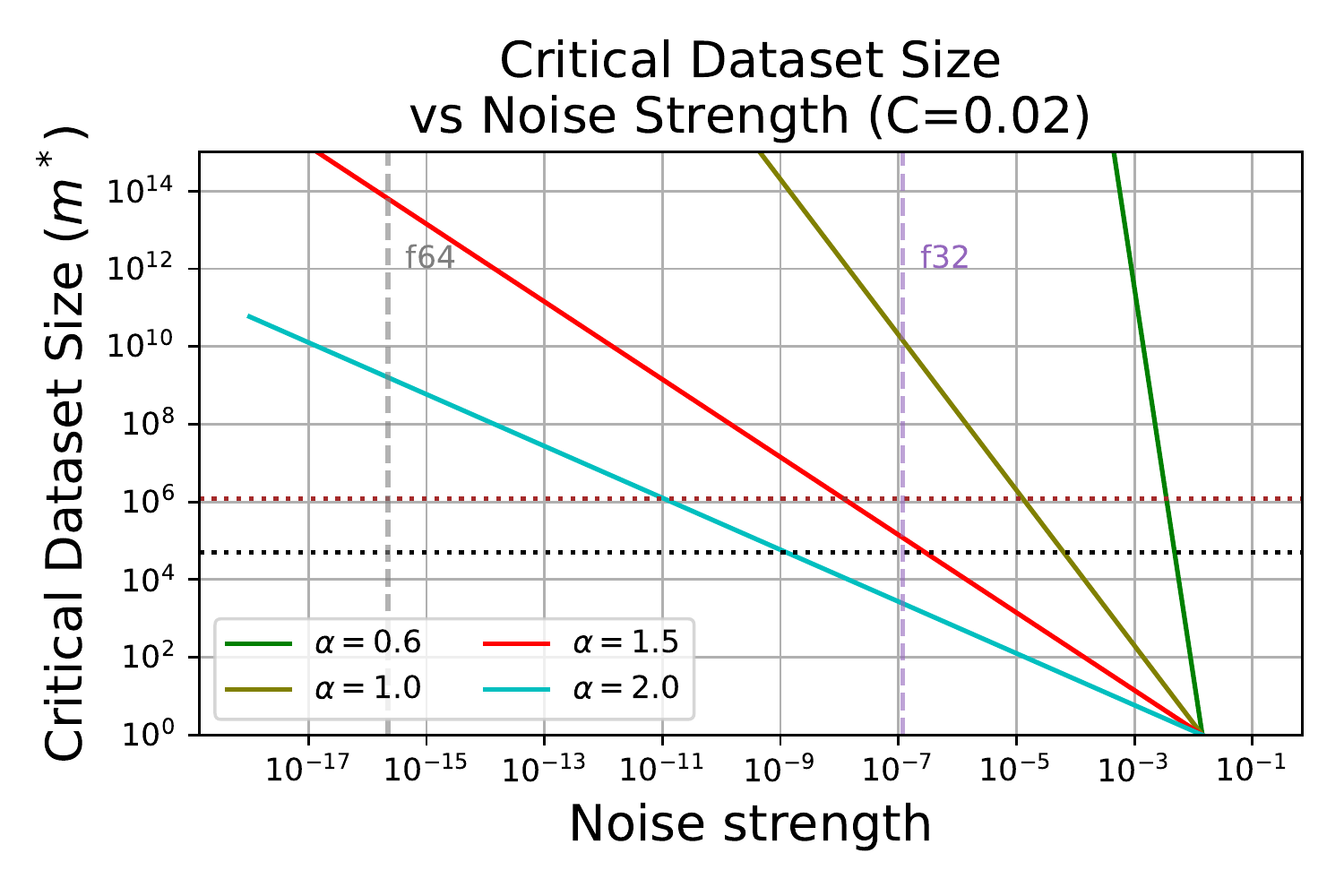}
 \put (0,0) {\textbf{\small(c)}}
 \end{overpic}
 
\caption{\textbf{The CNN-GAP architecture has poor
kernel conditioning} 
\textbf{(a)} Eigenvalue spectrum of infinite network kernels on 10k datapoints. Dashed lines are noise eigenvalue scale from \texttt{float32} precision. Eigenvalue for CNN-GAP's NNGP decays fast and negative eigenvalue may occur when dataset size is $O(10^4)$ in \texttt{float32} but is well-behaved with higher precision. 
\textbf{(b-c)} Critical dataset size as function of eigenvalue decay exponent $\alpha$ or noise strength $\sigma_n$ given by \eqref{eq:critical-m}. 
}
\label{app kernel spectra}
\end{figure}

\section{Data augmentation via kernel ensembling}
\label{app kernel ensembling}

We start considering general ensemble averaging of predictors. 
Consider a sequence of training sets $\{{\cal D}_i\}$ each consisting of $m$ input-output pairs $\{(x_1, y_1), \dots, (x_m, y_m)\}$ from a data-generating distribution. For a learning algorithm, which we use NNGP/NTK inference for this study, will give prediction $\mu (x^*, {\cal D}_i)$ of unseen test point $x^*$. It is possible to obtain better predictor by averaging output of different predictors
\begin{equation}
    \hat \mu(x^*) = \frac{1}{E}\sum_i^E \mu (x^*, {\cal D}_i)\, , 
\end{equation}
where $E$ denotes the cardinality of $\{{\cal D}_i\}$.  
This ensemble averaging is simple type of committee machine which has long history~\cite{clemen1989combining,dietterich2000ensemble}. While more sophisticated ensembling method exists (e.g.~\cite{freund1995desicion, breiman1996bagging,breiman2001random, opitz1996generating,opitz1999popular, rokach2010ensemble}), 
we strive for simplicity and considered naive averaging. One alternative we considered is generalizing average by
\begin{equation}
    \hat \mu_w(x^*) = \frac{1}{E}\sum_i^E w_i  \,\mu (x^*, {\cal D}_i)\,,
\end{equation}
were $w_i$ in general is set of weights satisfying $\sum_i w_i = 1$. We can utilize posterior variance $\sigma_i^2$ from NNGP or NTK with MSE loss via Inverse-variance weighting (IVW) where weights are given  as
\begin{equation}
    w_i = \frac{\sigma_i^{-2}}{\sum_j \sigma_j^{-2}} \,.
\end{equation}
In simple bagging setting~\cite{breiman1996bagging}, we observe small improvements with IVW over naive averaging. This indicates posterior variance for different draw of  $\{{\cal D}_i\}$ was quite similar. 

Application to data augmentation (DA) is simple as we consider process of generating $\{{\cal D}_i\}$ from a (stochastic) data augmentation transformation ${\cal T}$. We consider action of ${\cal T}(x, y) = T(x, y)$ be stochastic (e.g. $T$ is a random crop operator) with probability $p$ augmentation transformation (which itself could be stochastic) and probability $(1 - p)$ of $T = \text{Id}$. Considering ${\cal D}_0$ as clean un-augmented training set, we can imagine dataset generating process ${\cal D}_i \sim {\cal T} ({\cal D}_0)$, where we overloaded definition of ${\cal T}$ on training-set to be data generating distribution. 

\begin{figure}[h]
\centering
\includegraphics[width=\columnwidth]{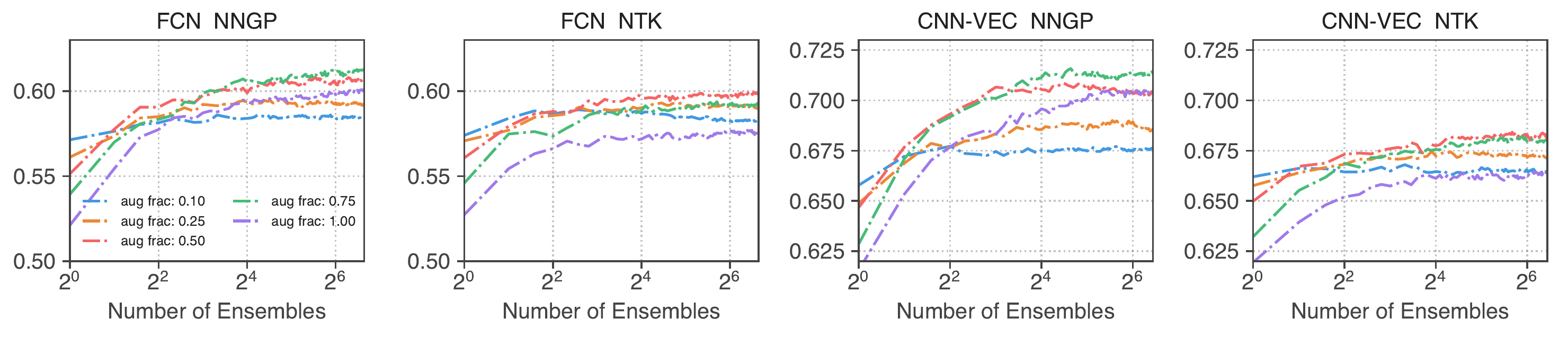}
\caption{\textbf{Data augmentation ensemble for infinite network kernels with varying augmentation fraction.} See also \sref{sec:data-augmentation}.}
\label{fig:kerne-da-ens-frac}
\end{figure}

For experiments in~\sref{sec:data-augmentation}, we took $T$ to be standard augmentation strategy of horizontal flip and random crop by 4-pixels with augmentation fraction $p = 0.5$ (see~\Figref{fig:kerne-da-ens-frac} for effect of augmentation fraction on kernel ensemble). In this framework, it is trivial to generalize the DA transformation to be quite general (e.g. learned augmentation strategy studied by~\citet{cubuk2019autoaugment, cubuk2019randaugment}).

\section{ZCA whitening}
\label{app ZCA}
Consider $m$ (flattened) $d$-dimensional training set inputs $X$ (a $d\times m$ matrix) with  data covariance
\begin{equation}
    \Sigma_X = \frac{1}{d} X X^T \,.
\end{equation}
The goal of whitening is to find a whitening transformation $W$, a $d\times d$ matrix, such that the features of transformed input 
\begin{equation}
    Y = W X
\end{equation}
are uncorrelated, e.g. $\Sigma_Y \equiv \frac 1 d YY^T= I$. Note that $\Sigma_X$ is constructed only from training set while $W$ is applied to both training set and test set inputs.
Whitening transformation can be efficiently computed by eigen-decomposition\footnote{For PSD matricies, it is numerically more reliable to obtain via SVD.}
\begin{equation}
    \Sigma_X = U D U^T\,
\end{equation}
where $D$ is diagonal matrix with eigenvalues, and $U$ contains eigenvector of $\Sigma_X$ as its columns. 

With this ZCA whitening transformation is obtained by following whitening matrix
\begin{align}
    W_\text{ZCA} &= U \sqrt{\left(D + \epsilon  \tfrac{tr(D)}{d} I_d\right)^{-1} } \, U^T \,.
\end{align}

Here, we introduced trivial reparameterization of conventional regularizer such that regularization strength $\epsilon$ is input scale invariant. It is easy to check $\epsilon \rightarrow 0$ corresponds to whitening with $\Sigma_Y = I$. In \sref{sec:zca}, we study the benefit of taking non-zero regularization strength for both kernels and finite networks. We denote transformation with non-zero regularizer, ZCA regularization preprocessing. ZCA transformation preserves spatial and chromatic structure of original image as illustrated in~\Figref{app ZCA}. Therefore image inputs are reshaped to have the same shape as original image. 

In practice, we standardize both training and test set per (RGB channel) features
of the training set before and after the ZCA whitening. This ensures transformed inputs are mean zero and variance of order 1.

\begin{figure}
\centering
  \begin{overpic}[width=\linewidth]{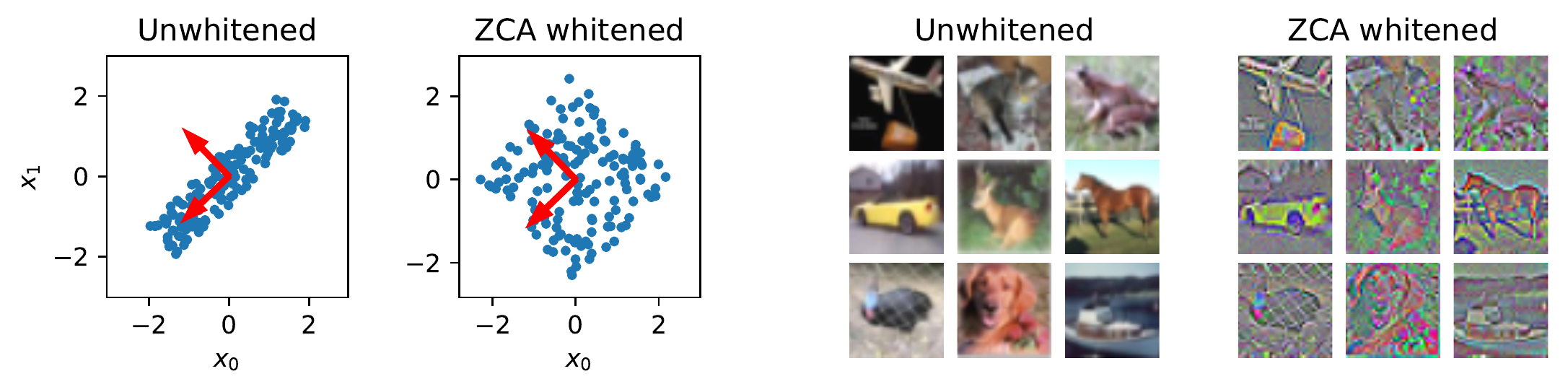}
 \put (0,1) {\textbf{\small(a)}}
 \put (48.5,1) {\textbf{\small(b)}}
\end{overpic}    
    \caption{
    \textbf{Illustration of ZCA whitening.} Whitening is a linear transformation of a dataset
    that removes correlations between feature dimensions, setting all non-zero eigenvalues of the covariance matrix to 1. 
    ZCA whitening is a specific choice of the linear transformation that rescales the data in the directions given by the eigenvectors of the covariance matrix, but without additional rotations or flips. 
    {\em (a)} A toy 2d dataset before and after ZCA whitening. Red arrows indicate the eigenvectors of the covariance matrix of the unwhitened data.
    {\em (b)} ZCA whitening of CIFAR-10 images preserves spatial and chromatic structure, while equalizing the variance across all feature directions. 
    Figure reproduced with permission from \citet{wadia2020whitening}. See also \sref{sec:zca}.
    }
\label{fig cifar zca}
\end{figure}

\section{MSE vs Softmax-cross-entropy loss training of neural networks}
\label{app:xent-vs-mse}
\begin{figure}[h]
\centering
\includegraphics[width=.4\columnwidth]{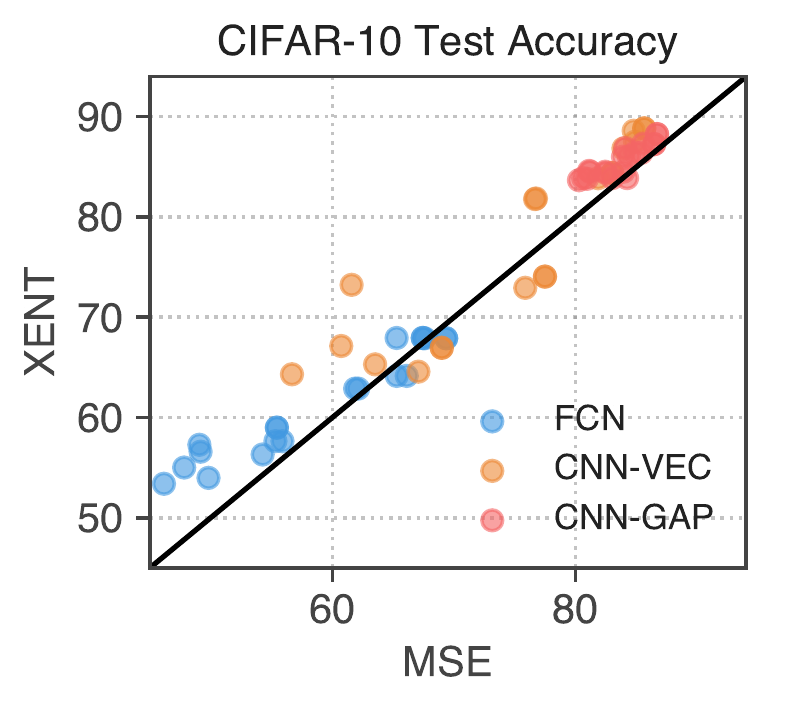}
\caption{\textbf{MSE trained networks are competitive while there is a clear benefit to using Cross-entropy loss}}
\label{fig:xent-vs-mse}
\end{figure}

Our focus was mainly on fininte networks trained with MSE loss for simple comparison with kernel methods that gives closed form solution. Here we present comparison of MSE vs softmax-cross-entropy trained networks. See Table~\ref{tab:xent-vs-mse} and \Figref{fig:xent-vs-mse}. 

\begin{table}
\centering
\caption{Effects of MSE vs softmax-cross-entropy loss on base networks with various interventions}\label{tab:xent-vs-mse}
\begin{tabular}{llllllll}
\toprule
Architecture &     Type & Param &   Base & +LR+U & +L2+U &  +L2+LR+U &   Best \\
\midrule \midrule
FCN &          MSE &   STD &  47.82  & 49.07 &  49.82 &  55.32 &  55.90 \\
  &            &   NTK &  46.16 & 49.17 & 54.27 &  55.44 &  55.44 \\
  &       XENT &   STD &  55.01 &  57.28 & 53.98 &  57.64 &  57.64 \\
  &            &   NTK &  53.39 &  56.59 & 56.31 &  58.99 &  58.99 \\
  &      MSE+DA &   STD &  65.29 &  66.11 & 65.28 &  67.43 &  67.43 \\
  &            &   NTK &  61.87 &  62.12 & 67.58 &  69.35 &  69.35 \\
  &    XENT+DA &   STD &  64.15 &  64.15 & 67.93 &  67.93 &  67.93 \\
  &            &   NTK &  62.88 &  62.88 &  67.90 &  67.90 &  67.90 \\
 \midrule
CNN-VEC &          MSE &   STD &  56.68 &  63.51 & 67.07 &  68.99 &  68.99 \\
  &           &   NTK &  60.73 &  61.58 & 75.85 &  77.47 &  77.47 \\
 &     XENT &   STD &  64.31 &  65.30 & 64.57 &  66.95 &  66.95 \\
 &      &   NTK &  67.13 &  73.23 & 72.93 &  74.05 &  74.05 \\
 &       MSE+DA &   STD &  76.73 & 81.84 &  76.66 &  83.01 &  83.01 \\
 &        &   NTK &  83.92 &  84.76 & 84.87 &  85.63 &  85.63 \\
 &   XENT+DA &   STD &  81.84 & 83.86 & 81.78 &  84.37 &  84.37 \\
 &           &   NTK &  86.83 & 88.59 & 87.49 &  88.83 &  88.83 \\
 \midrule
CNN-GAP &          MSE &   STD &  80.26 & 80.93 & 81.10 &  83.01 &  84.22 \\
 &           &   NTK &  80.61 & 82.44 & 81.17 &  82.43 &  83.92 \\
 &     XENT &   STD &  83.66 & 83.80 & 84.59 &  83.87 &  83.87 \\
 &        &   NTK &  83.87 &  84.40 & 84.51 &  84.51 &  84.51 \\
 &       MSE+DA &   STD &  84.36 &  83.88 & 84.89 &  86.45 &  86.45 \\
 &        &   NTK &  84.07 &  85.54 & 85.39 &  86.68 &  86.68 \\
 &  XENT+DA &   STD &  86.04 &  86.01 & 86.42 &  87.26 &  87.26 \\
 &   &   NTK &  86.87 &  87.31 & 86.39 &  88.26 &  88.26 \\
\bottomrule
\end{tabular}
\end{table}

\section{Comment on batch size}
\label{app:batch-size}
Correspondence between NTK and gradient descent training is direct in the full batch gradient descent (GD) setup (see~\cite{Dyer2020Asymptotics} for extensions to mini-batch SGD setting). Therefore base comparison between finite networks and kernels is the full batch setting. While it is possible to train our base models with GD, for full CIFAR-10 large emprical study becomes impractical. In practice, we use mini-batch SGD with batch-size $100$ for FCN and $40$ for CNNs. 

\begin{figure}
\centering
\begin{overpic}[width=0.8\linewidth]{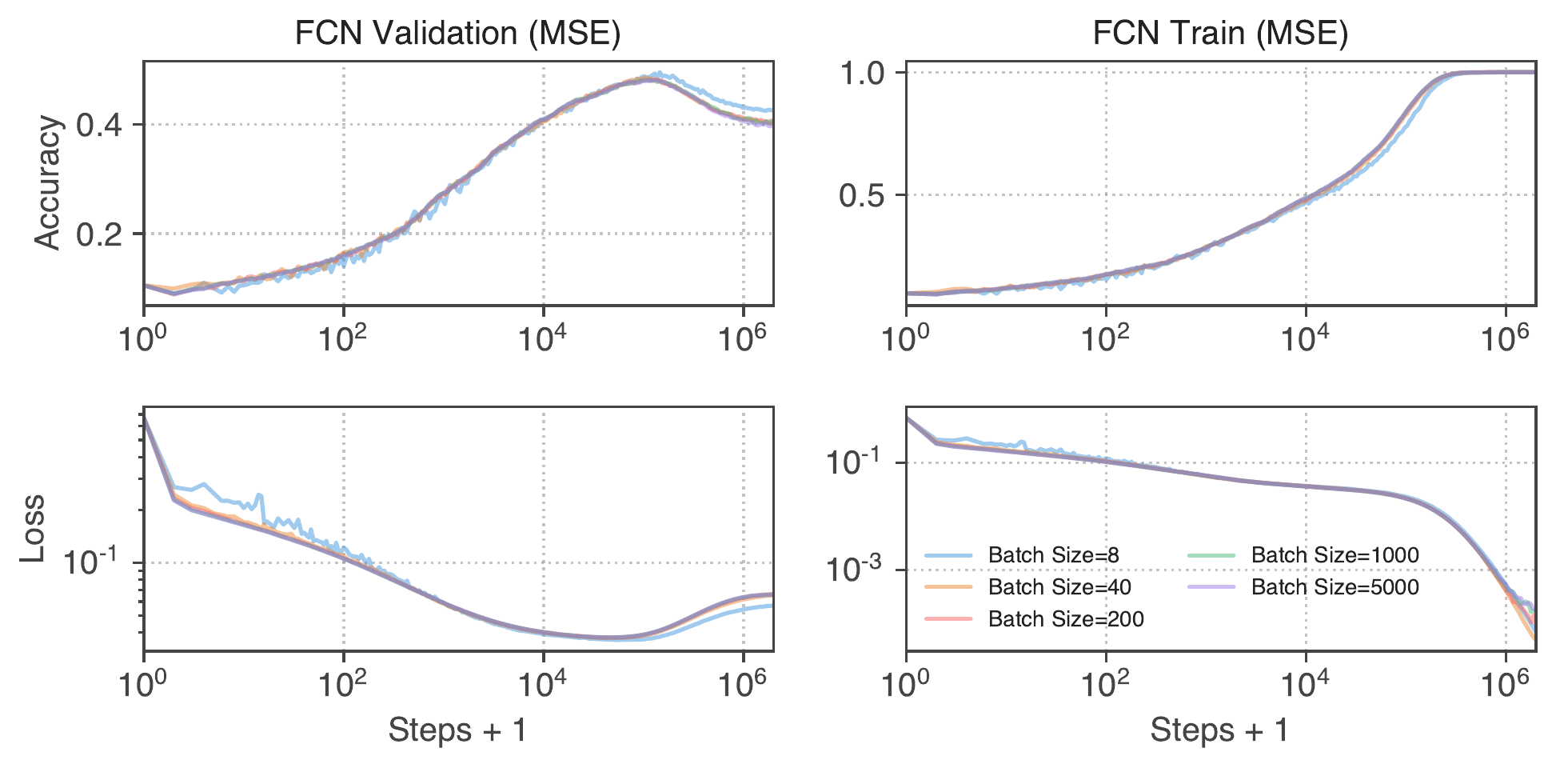}
 \put (0,0) {\textbf{\small(a)}}
\end{overpic}\\
\vspace{0.2cm}
\begin{overpic}[width=0.8\linewidth]{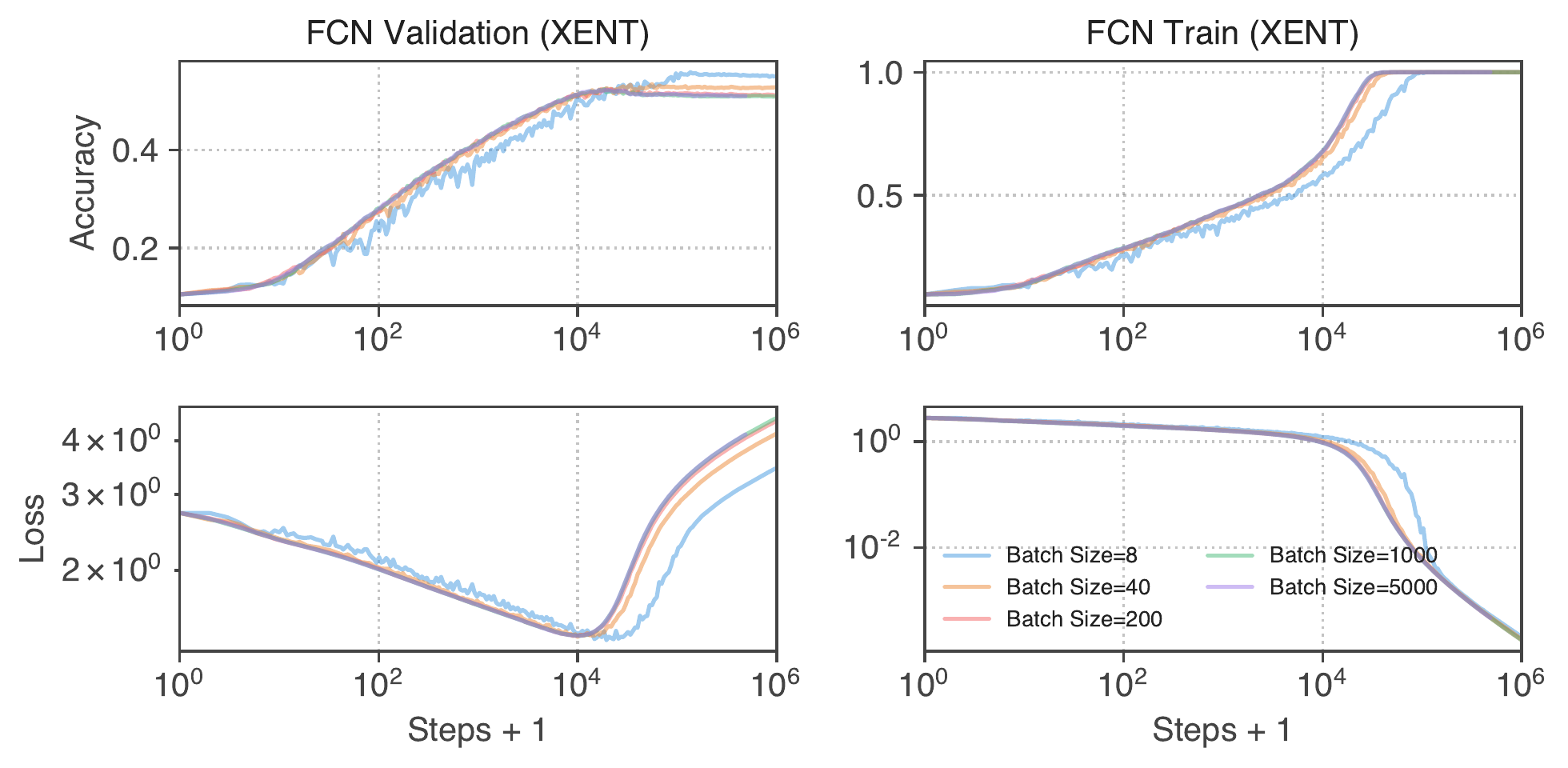}
 \put (0,0) {\textbf{\small(b)}}
\end{overpic}
\\
\vspace{0.2cm}
\begin{overpic}[width=0.8\linewidth]{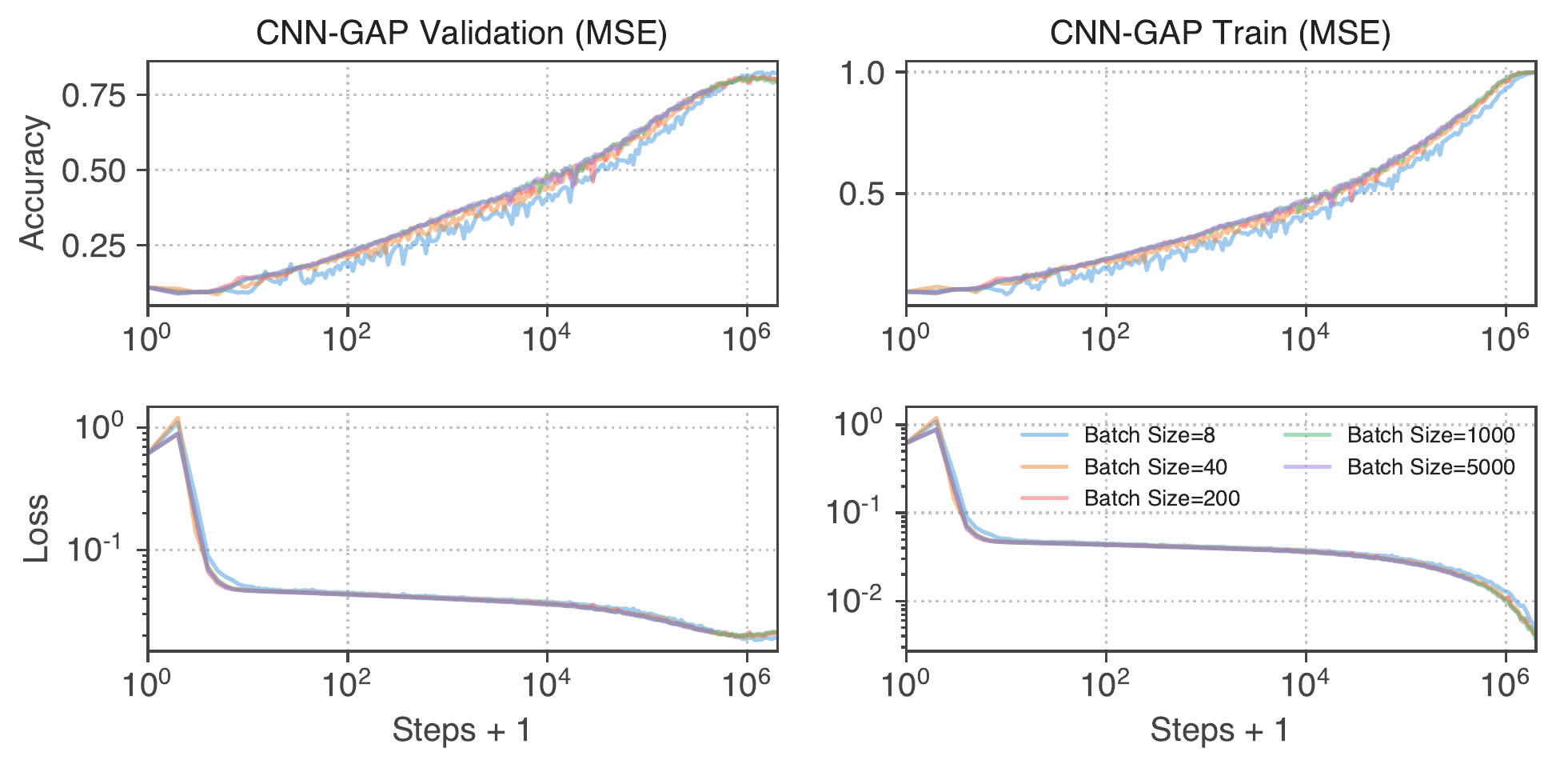}
 \put (0,0) {\textbf{\small(c)}}
\end{overpic}\\ 
\vspace{0.2cm}
\begin{overpic}[width=0.8\linewidth]{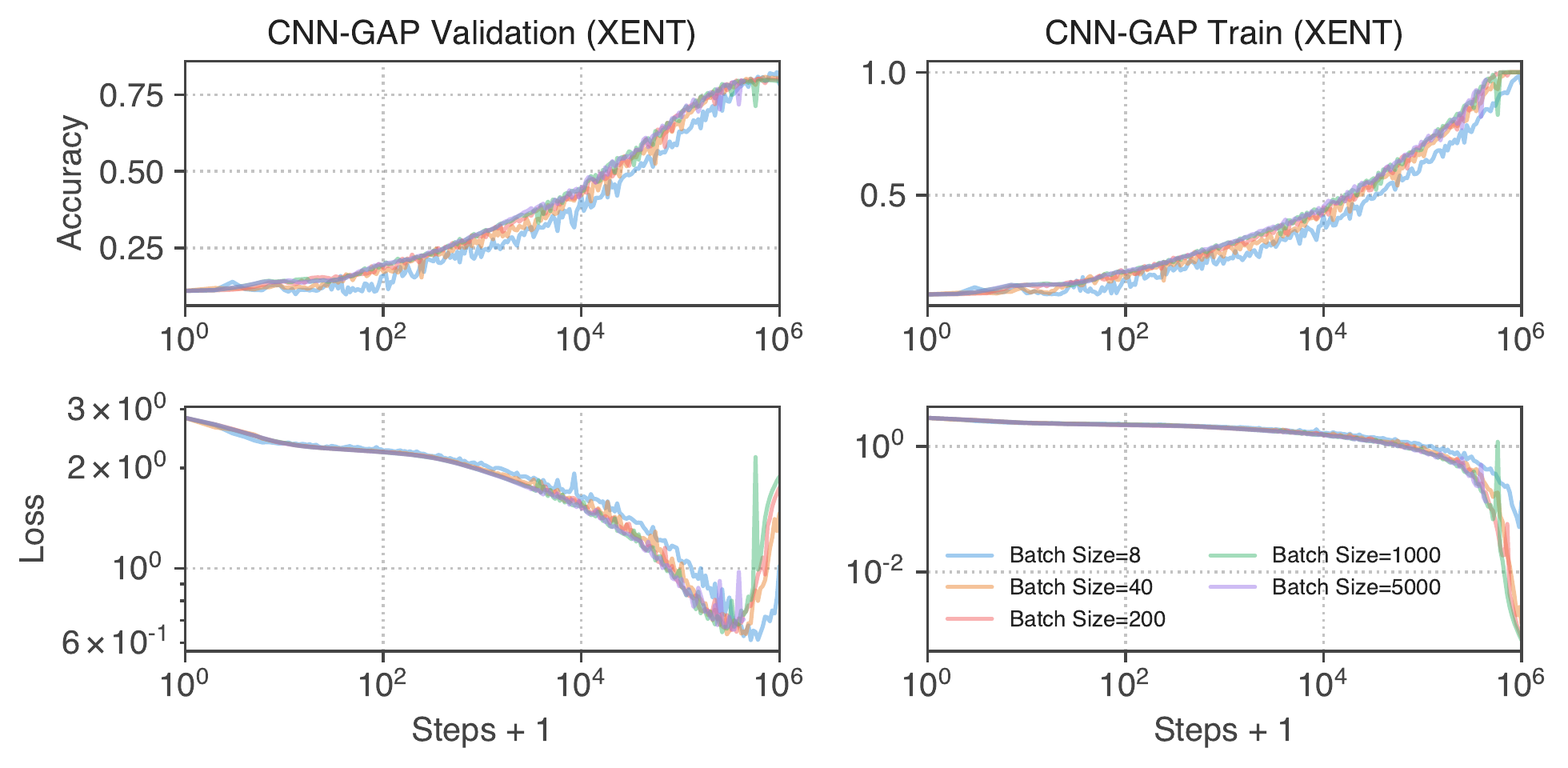}
 \put (0,0) {\textbf{\small(d)}}
\end{overpic}

    \caption{\textbf{Batch size does not affect training dynamics for moderately large batch size.}}
    \label{fig:bs}
\end{figure}

We studied batch size effect of training dynamics in \Figref{fig:bs} and found that for these batch-size choices does not affecting training dynamics compared to much larger batch size. 
\citet{shallue2018measuring, mccandlish2018empirical} observed that universally for wide variety of deep learning models  there are batch size beyond which one could gain training speed benefit in number of steps. We observe that maximal useful batch-size in workloads we study is quite small.

\section{Addtional tables and plots}

\begin{table}[h]
\centering
\caption{\textbf{CIFAR-10 classification mean squared error(MSE) for nonlinear and linearized finite neural networks, as well as for NTK and NNGP kernel methods}.
Starting from \texttt{Base} network of given architecture class described in \sref{sec:experimental_design}, performance change of \textbf{centering} (\texttt{+C}), \textbf{large learning rate} (\texttt{+LR}), allowing \textbf{underfitting} by early stopping (\texttt{+U}), input preprocessing with \textbf{ZCA regularization} (\texttt{+ZCA}), multiple initialization \textbf{ensembling} (\texttt{+Ens}), and some combinations are shown, for {\color{standard_param}\textbf{Standard}} and {\color{ntk_param}\textbf{NTK}} parameterization. See also Table~\ref{tab:main-table} and \Figref{fig:tricks_vs_accuracy} for accuracy comparison.}
\vspace{0.3cm}
\label{tab:main-table-mse}
\resizebox{\columnwidth}{!}{%

\begin{tabular}{@{}lc|ccccccccc|cc|cc@{}}
\toprule
{} &
  Param &
  Base &
  +C &
  +LR &
  +L2 &
  \begin{tabular}[c]{@{}l@{}}+L2 \\ +U\end{tabular}
 &
  \begin{tabular}[c]{@{}c@{}}+L2 \\ +LR\end{tabular} &
  \begin{tabular}[c]{@{}l@{}}+L2 \\+LR\\ +U \end{tabular} &
  +ZCA &
  \begin{tabular}[c]{@{}c@{}} Best \\ w/o DA\end{tabular} &
  +Ens &
  \begin{tabular}[c]{@{}l@{}}+Ens \\+C \end{tabular}
&
 \begin{tabular}[c]{@{}l@{}} +DA\\ +U \end{tabular}&
  \begin{tabular}[c]{@{}l@{}}+DA \\+L2\\ +LR\\ +U\end{tabular} \\

\midrule \midrule
      FCN &   STD &  0.0443 &  0.0363 &  0.0406 &  0.0411 &  0.0355 &  0.0337 &  0.0329 &  0.0483 &  0.0319 &   0.0301   &  0.0304     &0.0267 &  0.0242 \\
  &   NTK &  0.0465 &  0.0371 &  0.0423 &  0.0338 &  0.0336 &  0.0308 &  0.0308 &  0.0484 &  0.0308 &    0.0300  &  0.0302     &  0.0281 &  0.0225 \\
  \midrule
     CNN-VEC &   STD &  0.0381 &  0.0330 &  0.0340 &  0.0377 &  0.0279 &  0.0340 &  0.0265 &  0.0383 &  0.0265 &   0.0278   &  0.0287     &  0.0228 &  0.0183 \\
 &   NTK &  0.0355 &  0.0353 &  0.0355 &  0.0355 &  0.0231 &  0.0246 &  0.0227 &  0.0361 &  0.0227 &   0.0254    &    0.0278   & 0.0164 &  0.0143 \\
 \midrule
    CNN-GAP &   STD &  0.0209 &  0.0201 &  0.0207 &  0.0201 &  0.0201 &  0.0179 &  0.0177 &  0.0190 &  0.0159 & 0.0172     &    0.0165   &  0.0185 &  0.0149 \\
 &   NTK &  0.0209 &  0.0201 &  0.0195 &  0.0205 &  0.0181 &  0.0175 &  0.0170 &  0.0194 &  0.0161 & 0.0163     &   0.0157    &  0.0186 &  0.0145 \\
\bottomrule
\end{tabular}%
}
\\
\vspace{0.2cm}

\resizebox{\columnwidth}{!}{%
\begin{tabular}{@{}lc|cccccc||ccc|ccc@{}}
\toprule
 &
  Param &
  Lin Base &
  +C &
  +L2 &
  \begin{tabular}[c]{@{}l@{}}+L2 \\+U \end{tabular} &
  +Ens &
  \begin{tabular}[c]{@{}l@{}}+Ens \\+C \end{tabular}&
  NTK & +ZCA & \begin{tabular}[c]{@{}c@{}}+DA \\ +ZCA\end{tabular}&
  NNGP &+ZCA &
\begin{tabular}[c]{@{}c@{}}+DA \\ +ZCA\end{tabular}\\
  \midrule\midrule
  FCN &   STD &  0.0524 &  0.0371 &  0.0508 &  0.0350 & 0.0309 & 0.0305 & 0.0306 & 0.0302 & - & \multirow{2}*{0.0309} & \multirow{2}*{0.0308} & \multirow{2}*{0.0297}  \\
 &   NTK &  0.0399 &  0.0366 &  0.0370 &  0.0368 &  0.0305 & 0.0304 &0.0305 & 0.0302 & 0.0298 \\
 \midrule
  CNN-VEC &   STD &  0.0436 &  0.0322 &  0.0351 &  0.0351 & 0.0293 & 0.0291 & 0.0287 & 0.0277 & - & \multirow{2}*{0.0286} & \multirow{2}*{0.0281} & \multirow{2}*{0.0256}\\
 &   NTK &  0.0362 &  0.0337 &  0.0342 &  0.0339 &  0.0286 &0.0286 & 0.0283 & 0.0274 & 0.0273 \\
 \midrule
  CNN-GAP &   STD &  \multicolumn{6}{c||}{< 0.0272* (Train accuracy 86.22 after 14M steps)} &  0.0233 & 0.0200& - & \multirow{2}*{0.0231} & \multirow{2}*{0.0204} & \multirow{2}*{0.0191} \\
 &   NTK &  \multicolumn{6}{c||}{< 0.0276* (Train accuracy 79.90 after 14M steps)}  & 0.0232 & 0.0200& 0.0195\\
 \bottomrule
\end{tabular}%
}
\end{table}

\begin{figure}[h]
\centering
\includegraphics[width=\columnwidth]{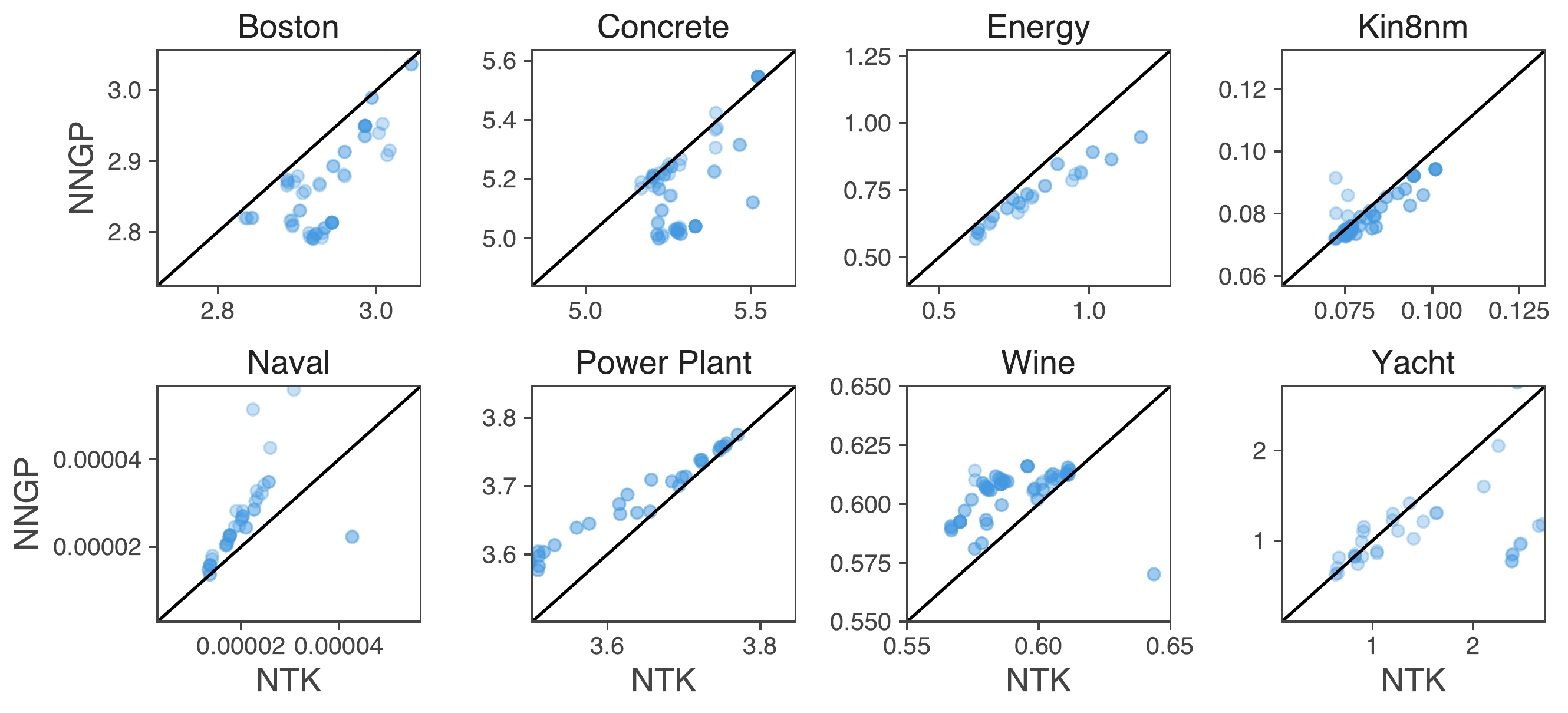}
\caption{\textbf{On UCI dataset NNGP often outperforms NTK on RMSE.}
We evaluate predictive performance of FC NNGP and NTK on UCI regression dataset in the standard 20-fold splits first utilized in~\cite{hernandez2015probabilistic, gal2016dropout}. We plot average RMSE across the splits. Different scatter points are varying hyperparameter settings of (depth, weight variance, bias variance). In the tabular data setting, dominance of NNGP is not as prominent across varying dataset as in image classification domain. 
}
\label{fig:nngp-vs-ntk-uci}
\end{figure}

\begin{figure}[h]
\centering
\includegraphics[width=\columnwidth]{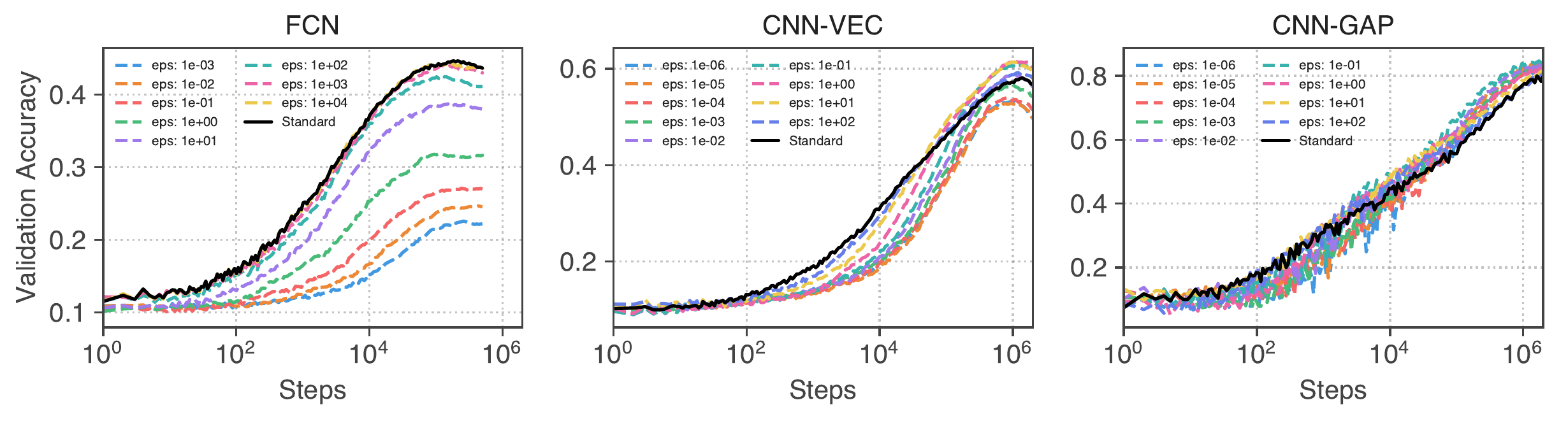}
\includegraphics[width=\columnwidth]{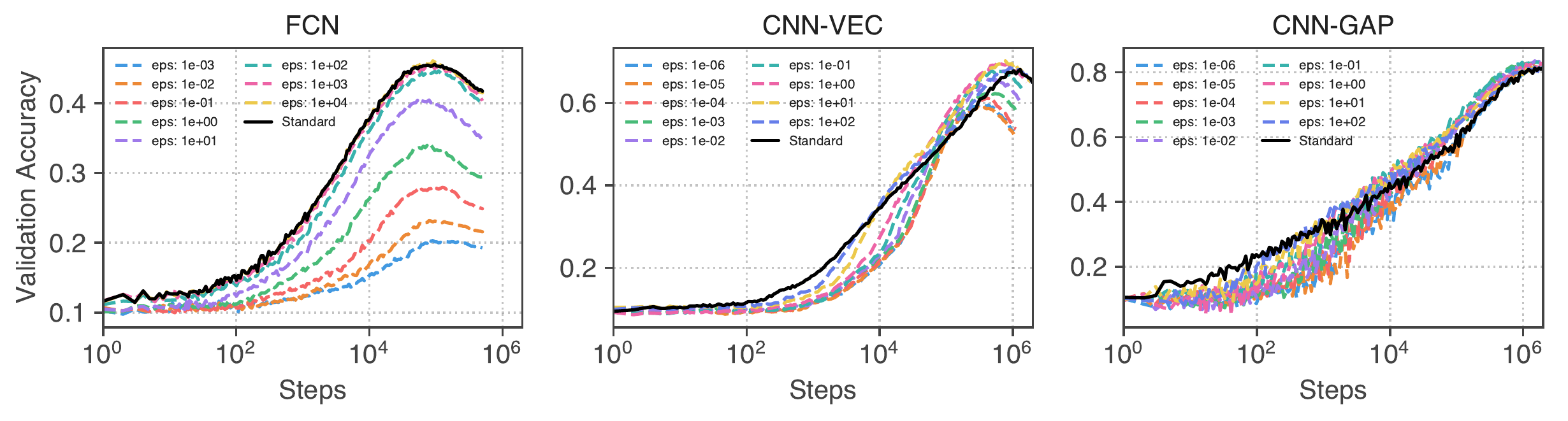}
\caption{\textbf{ZCA regularization helps finite network training.}
(\textbf{upper}) Standard parameterization, (\textbf{lower}) NTK parameterization. See also \sref{sec:zca} and \Figref{fig:zca}.
}
\label{fig:app-zca-training}
\end{figure}

\end{document}